\documentclass{article}

    \usepackage[final,nonatbib]{neurips_2023}

\usepackage[utf8]{inputenc} % allow utf-8 input
\usepackage[T1]{fontenc}    % use 8-bit T1 fonts
\usepackage{hyperref}       % hyperlinks
\usepackage{url}            % simple URL typesetting
\usepackage{booktabs}       % professional-quality tables
\usepackage{amsfonts}       % blackboard math symbols
\usepackage{nicefrac}       % compact symbols for 1/2, etc.
\usepackage{microtype}      % microtypography
\usepackage{amsmath}
\usepackage{amssymb}

\usepackage{graphicx}
\usepackage[capitalize]{cleveref}
\crefname{section}{Sec.}{Secs.}
\Crefname{section}{Section}{Sections}
\Crefname{table}{Table}{Tables}
\crefname{table}{Tab.}{Tabs.}

\usepackage{caption}
\usepackage{subcaption}

\usepackage{multirow}

\usepackage{pifont}
\newcommand{\cmark}{\ding{51}}
\newcommand{\xmark}{\ding{55}}

\usepackage[dvipsnames]{xcolor}

\def\supp{\textit{\textcolor{BrickRed}{supplementary materials}}}

\def\ddd{D$^3$}

\newlength{\twosubht}
\newsavebox{\twosubbox}

\title{Described Object Detection: Liberating Object Detection with Flexible Expressions}

\author{%
  Chi~Xie$^1$\footnotemark[2] \quad
  Zhao~Zhang$^{2}$\footnotemark[2] \quad
  Yixuan~Wu$^3$ \quad
  Feng~Zhu$^2$ \quad
  Rui~Zhao$^{2}$ \quad
  Shuang~Liang$^{1}$\footnotemark[1] \quad
  \\
  $^{1}${Tongji University} \quad
  $^{2}${Sensetime Research} \quad
  $^{3}${Zhejiang University} \quad \\
  {\tt\small chixie@tongji.edu.cn \quad zzhang@mail.nankai.edu.cn \quad shuangliang@tongji.edu.cn}
}

\begin{document}

\maketitle

\renewcommand{\thefootnote}{\fnsymbol{footnote}}
\footnotetext[1]{Corresponding author.}
\footnotetext[2]{Equal contribution.}

\begin{abstract}
    Detecting objects based on language information is a popular task that includes Open-Vocabulary object Detection (OVD) and Referring Expression Comprehension (REC). In this paper, we advance them to a more practical setting called \textit{Described Object Detection} (DOD) by expanding category names to flexible language expressions for OVD and overcoming the limitation of REC only grounding the pre-existing object. 
    We establish the research foundation for DOD by constructing a \textit{Description Detection Dataset} (\ddd{}). This dataset features flexible language expressions, whether short category names or long descriptions, and annotating all described objects on all images without omission.
    By evaluating previous SOTA methods on \ddd{}, we find some troublemakers that fail current REC, OVD, and bi-functional methods.
    REC methods struggle with confidence scores, rejecting negative instances, and multi-target scenarios, while OVD methods face constraints with long and complex descriptions. Recent bi-functional methods also do not work well on DOD due to their separated training procedures and inference strategies for REC and OVD tasks.
    Building upon the aforementioned findings, we propose a baseline that largely improves REC methods by reconstructing the training data and introducing a binary classification sub-task, outperforming existing methods.
    Data and code are available at \href{https://github.com/shikras/d-cube}{this URL} and related works are tracked in \href{https://github.com/Charles-Xie/awesome-described-object-detection}{this repo}.
\end{abstract}
\section{Introduction}
\label{sec:introduction}

\begin{figure}[t]
\begin{center}
   \includegraphics[width=\linewidth]{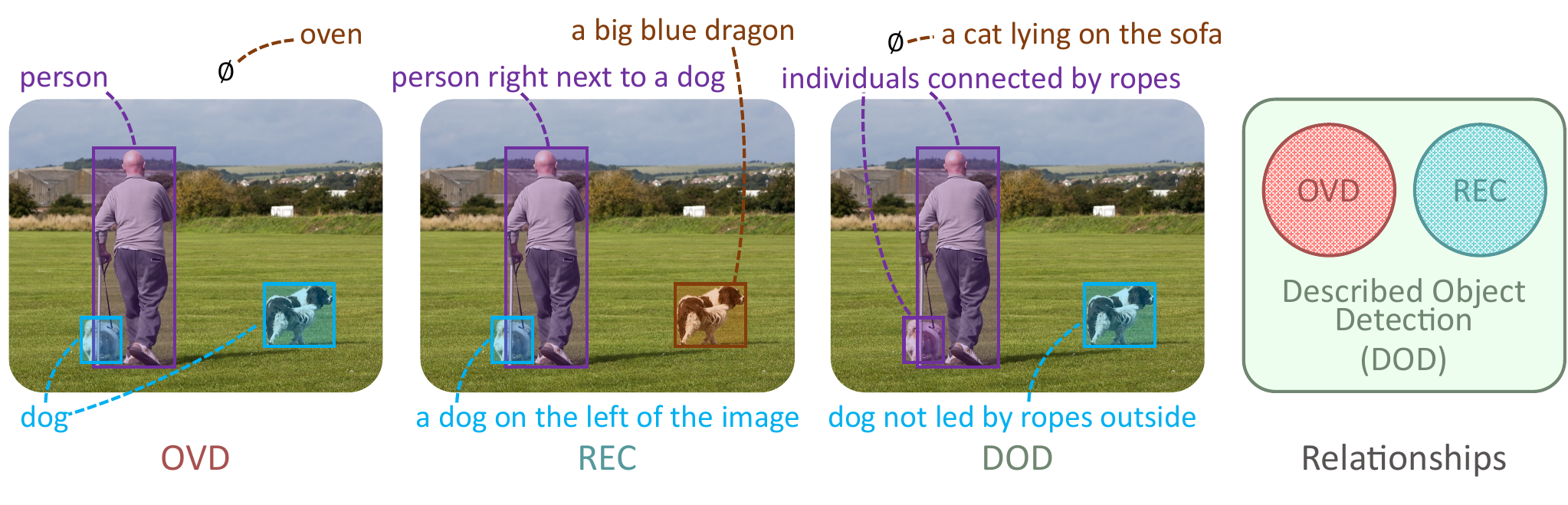}
    % \begin{subfigure}[b]{0.32\textwidth}
    %     \centering
    %     \captionsetup{justification=centering}
    %     \includegraphics[width=\textwidth]{figures/teaser/teaser_ovd.pdf}
    %     \caption{(open-vocabulary) \\\hspace{\textwidth} object detection}
    %     \label{fig:teaser_ovd}
    % \end{subfigure}
    % \hfill
    % \begin{subfigure}[b]{0.32\textwidth}
    %     \centering
    %     \captionsetup{justification=centering}
    %     \includegraphics[width=\textwidth]{figures/teaser/teaser_rec.pdf}
    %     \caption{referring expression \\\hspace{\textwidth} comprehension}
    %     \label{fig:teaser_rec}
    % \end{subfigure}
    % \hfill
    % \begin{subfigure}[b]{0.32\textwidth}
    %     \centering
    %     \captionsetup{justification=centering}
    %     \includegraphics[width=\textwidth]{figures/teaser/teaser_ours.pdf}
    %     \caption{omni-expression \\\hspace{\textwidth} detection}
    %     \label{fig:teaser_ours}
    %     \end{subfigure}
    \vspace{-25pt}
\end{center}
   \caption{Examples showing the difference between REC, OVD and Described Object Detection (DOD).
   OVD detects arbitrary number (including zero, denoted with $\emptyset$) of objects based on a category name; REC grounds one region based on a language description, whether the object truly exists or not; DOD detect all instances on each image in the dataset, based on a flexible reference.
   % \Todo{a + b != c.}
   \vspace{-8pt}
   }
\label{fig:teaser}
\end{figure}

Detecting objects of interest within a scene using language is a pivotal area of focus. This field encompasses two key tasks: Open-Vocabulary object Detection (OVD)~\cite{ghiasi2022scaling,gu2022openvocabulary,li2022grounded,minderer2022simpleOWLViT,zang2022open,zareian2021open} and Referring Expression Comprehension (REC)~\cite{li2021referring,luo2020multi,liu2023polyformer,yu2016modeling,zhou2021real}.
We present an intuitive illustration of these two settings in \cref{fig:teaser}.
The first task, OVD, expands the scope of object detection (OD) to any given short category name. However, these settings neglect the instances described by intricate descriptions.
The second task, REC, focuses on spatially locating one target described by an expression and assumes the target must exist in the image. However, in real-world scenarios, if the described objects do not exist in the image, REC algorithms output false-positive results.
Recent advancements have witnessed the joint training of bi-functional models, such as Grounding-DINO~\cite{liu2023groundingdino} and UNINEXT~\cite{yan2023universal}, which involve both OVD and REC data.
Notwithstanding, these models still rely on separate training procedures and inference strategies for OVD and REC, and evaluate these two tasks independently.

As shown in \cref{fig:teaser}, a more practical detection algorithm should be able to detect any described category, whether long or short, complex or simple, while discarding predictions in images where targets are absent.
In order to address this significant yet often overlooked scenario, we propose the concept of \textbf{Described Object Detection (DOD)}. Note that this setting is a superset of OVD and REC.
When the language expression is limited to a short category name, it becomes OVD. When we limit the images to detect objects known to be present in the images beforehand, it downgrades to REC.

Can the existing SOTA algorithms of the community support DOD tasks?
To address this inquiry, this paper establishes the research foundation of DOD tasks by constructing a dataset, scrutinizing relevant methodologies, analyzing the relevant methods, and exploring improvement space.

\textbf{Motivation \& real-world application of DOD.}
OVD is limited to categorical detection, focusing on \textit{classes} rather than specific attributes or relationships. It lacks detailed contextual understanding and cannot adapt to precise detection requirements from language.
REC comprehend longer descriptions for attributes or relationships, but assumes the existence of one target in the image. This leads to false positives when the target is absent, limiting its practical usability.
Consider detecting \texttt{individuals without helmets} on a construction site using camera data: OVD can detect \texttt{helmets} and \texttt{people} but not determine their relationship. REC locate one region in any image and generate false positives frequently. Existing solutions involve using separate models for object detection then relationship classification, or REC after image classification, both resulting in inefficiency.

% Hence, there is a significant demand for detection based on language descriptions: a model with strong generalization capabilities, capable of determining whether the described object exists in the image and localizing it based on arbitrary language descriptions. This is where our proposed DOD task comes in. The introduced DOD task has various practical applications, including:
% Urban security, like detecting ``individuals without helmets'' in construction sites, ``dog outside without leash'' in communities, ``clothes hung outdoors'' on a street, ``overloaded vehicles'' and ``fallen trees on roadsides'' on the road, etc;
% Network security, where sensitive images containing bloodshed or violence need to be detected within a massive image dataset;
% (Fine-grained) photo album retrieval based on language (descriptions, keywords, etc.);
% Retrieval and filtering of web image data;
% Detection of specific events in autonomous driving, such as ``pedestrians crossing the road''.
Hence, there is a demand for language-based object detection: a model with strong generalization capabilities that can verify the existence of described objects in images and localize them based on arbitrary expressions. The proposed DOD task addresses this need and finds practical applications in:
urban security, detecting
% \texttt{individuals without helmets} in construction sites,
\texttt{dogs without leashes} in communities, \texttt{clothes hung outdoors} on streets, \texttt{overloaded vehicles}, and \texttt{fallen trees on roadsides};
network security, like identifying sensitive images with violence or bloodshed within large datasets;
(fine-grained) photo album retrieval based on descriptions or keywords;
retrieval and filtering of web image data;
specific event detection in autonomous driving, such as \texttt{pedestrians crossing the road}.

\textbf{Dataset \& benchmark.}
% For the DOD task, we built a new dataset called \textbf{Description Detection Dataset} (\ddd{}, /dikju:b/), comprising 422 well-designed descriptions and 24,282 positive object-description pairs. 
% Compared to the former OVD or REC datasets, as depicted in (\cref{fig:highlight}), it has three noteworthy characteristics:
% 1) All objects referred by descriptions are annotated across the entire dataset, making \ddd{} a detection-style dataset similar to COCO~\cite{lin2014microsoft}, rather than a REC dataset.
% 2) Instances in this dataset are annotated with flexible and free-form language expressions, which can be short or long, simple or complex, unlike the OVD dataset.
% 3) We included numerous absence descriptions, such as ``a person without a safety helmet'', addressing a widely-needed yet overlooked detection requirement. This is summarized in \cref{tab:dataset_highlight}.
% On \ddd{}, we evaluate three types of SOTA methods, namely OWL-ViT~\cite{minderer2022simpleOWLViT}/CORA~\cite{wu2023cora}(OVD), OFA (REC)~\cite{wang2022ofa}, and UNINEXT~\cite{yan2023universal}/Grounding-DINO~\cite{liu2023groundingdino} (bi-functional methods). This serves as a reference for the community.
For DOD, we introduce the \textbf{Description Detection Dataset} (\ddd{}, /dikju:b/), an evaluation-only benchmark containing 422 descriptions and 24,282 positive object-description pairs. Unlike previous OVD or REC datasets (see \cref{fig:highlight}), \ddd{} stands out in three key aspects (see \cref{tab:dataset_highlight}):
1) \textit{Complete annotation}: All descriptions refer to objects annotated throughout the dataset, making \ddd{} a detection-style dataset akin to COCO~\cite{lin2014microsoft}.
2) \textit{Unrestricted description}: Annotations in \ddd{} include diverse and flexible language expressions, varying in length and complexity.
3) \textit{Absence expression}: We include descriptions regarding absence of concepts, such as \texttt{a person \textit{without} a safety helmet}, addressing an often-overlooked detection requirement.
The details of \ddd{} is elaborated in \cref{sec:dataset}.
We evaluate state-of-the-art methods on \ddd{}: OWL-ViT~\cite{minderer2022simpleOWLViT}/CORA~\cite{wu2023cora} (OVD), OFA (REC)~\cite{wang2022ofa}, and UNINEXT~\cite{yan2023universal}/Grounding-DINO~\cite{liu2023groundingdino} (bi-functional) to provide a reference for the community. This benchmark may serve as a starting point for the DOD task.

\textbf{Findings \& improvements.} 
The experimental analysis for different methods on \ddd{} yields some findings for future research (see \cref{sec:experiments}):
1) Existing REC methods perform poorly, lacking confidence scores and the ability to reject negatives, and struggling with multi-target situations. This is due to their task formulation of grounding, i.e., matching between text and image region and not distinguishing positive and negatives.
2) OVD methods excel REC ones on DOD, though lengthy descriptions, which is not available in their training data, limit their performance.
3) Bi-functional methods, while superior to REC and OVD ones, share similar challenges with REC methods. Sometimes they are surpassed by OVD models, indicating they have not fully benefited from REC and OVD.
Based on these findings, we propose \textbf{a baseline OFA-DOD} that greatly improves a REC method, and outperforms current SOTAs. Its abilities to handle multiple targets and reject negative instances are improved by simple data reconstruction and an auxiliary sub-task. It is still far from a strong DOD method, but may provide some insights for research in the future.

\begin{table}[t]
    \centering
    \caption{Comparison between the proposed dataset and previous REC datasets and OVD datasets.}
    \begin{tabular}{l | c c c c}
    \hline
    \multirow{2}{*}{Dataset} & annotation & unrestricted & absence & instance-level \\
    & completeness & description & expression & annotation \\
    \hline
    RefCOCO & image-wise & \cmark & \xmark & \cmark \\
    COCO & dataset-wise & \xmark & \xmark & \cmark \\
    GRD & group-wise & \cmark & \xmark & \xmark \\
    Ours & dataset-wise & \cmark & \cmark & \cmark \\
    \hline
    \end{tabular}
    % \vspace{-15pt}
    \label{tab:dataset_highlight}
\end{table}

% The dataset is built by re-annotation on a previous dataset GRD, which is proposed for segmentation tasks like referring expression segmentation (RES). Our effort for extension over GRD covers the latter 2 characteristics, i.e., complete annotation across the dataset and absence expressions of objects. Besides, GRD itself only contains semantic-level annotation, and we further provide \textit{instance-level annotation} on it.

% Our dataset for omni-expression detection, build on the generalization of two different types of object understanding tasks (language-guided grounding and category-guided detection), provides a benchmark to verify the capabilities of models to discover and locate complex events. It is very challenging.

% We verify multiple state-of-the-art methods for REC, OVD or their mixture on the proposed benchmark, and propose a new method improved upon OFA. It outperforms existing methods, and can serve as a baseline for future works on language-guided detection.

% The contribution of this work includes:
% \begin{itemize}
%     \item Based on the limitation of existing object understanding tasks, we propose a more realistic and flexible setting called omni-expression detection;
%     \item We proposed a benchmark for evaluation under this setting, which has 4 characteristics that differs from previous benchmarks;
%     \Todo{rewrite this.}
%     \item We evaluate SOTA methods of different object understanding tasks on the proposed dataset, together with a improved model that outperforms them and serve as a proposed baseline.
% \end{itemize}

\section{Related Work}
\vspace{-5pt}
\label{sec:related_work}
\subsection{Relevant datasets and benchmarks}
\vspace{-5pt}
\noindent \textbf{Object detection datasets.}
% Object detection is a mature task in computer vision and many datasets have been proposed for this task.
% PASCAL VOC, MSCOCO, LVIS, Object365, OpenImage, ODinW, BigDetection, V3Det.
A variety of datasets have been proposed for object detection. Some have become standard benchmarks, like PASCAL VOC~\cite{Everingham2010} and COCO~\cite{lin2014microsoft}; while others are more frequently used for pretraining~\cite{shao2019objects365,Krasin2017,cai2022bigdetection}. A few works have focused on special settings, such as LVIS~\cite{gupta2019lvis} for long-tailed detection and ODinW~\cite{li2022elevater} for zero-shot evaluation in the wild.
Recently, V3Det~\cite{wang2023v3det}facilitates object detection with an extremely large vocabulary.
Some are re-splitted and frequently used in OVD as well, like COCO and LVIS.
As explained in \cref{sec:introduction}, these datasets are all annotated with simple category labels rather than flexible language expressions like \ddd{}.

\noindent \textbf{Referring expression comprehension datasets.}
% Referece expression comprehension is a multi-modal task.
% RefCOCO, RefCOCOg, RefCOCO+. RefClef. Visual Genome. PhraseCut.
% \Todo{notice that phraseCut refers to multiple instances per image.}
% Referring expression comprehension (REC) aims to localize an object in the given image described by a language expression.
Several datasets have been introduced to evaluate REC methods, including RefClef~\cite{kazemzadeh2014referitgame}, RefCOCO~\cite{yu2016modeling}, RefCOCO+~\cite{yu2016modeling}, RefCOCOg~\cite{mao2016generation}, Visual Genome~\cite{krishna2017visualgenome}, and PhraseCut~\cite{wu2020phrasecut}.
Some~\cite{kazemzadeh2014referitgame,yu2016modeling} are collected interactively, and the expressions are more concise and less diverse.
% A characteristic of RefCOCO+ is that location words are banned in its expressions, which also makes it more challenging.
% Among them, RefCOCO+ bans location words in expressions, making it more challenging.
RefCOCOg is collected non-interactively, resulting in more complex expressions.
Comparatively, Visual Genome focuses on visual relationships.
All these datasets only annotate a few positive images for each category and leave other images unknown, which makes them unsuitable for the detection task.
% One of the key strengths of Visual Genome lies in its extensive annotations of visual relationships, which enables researchers to investigate complex scene understanding, contextual reasoning, and the interplay between objects in an image.
% PhraseCut, collected on top of the Visual Genome, where each image is accompanied by a set of annotated phrases that describe the objects present in the scene. These phrases may include object attributes, relationships, or other linguistic descriptions that provide a more comprehensive understanding of the visual content.

\noindent \textbf{Other related tasks and datasets.}
Several related tasks and benchmarks exist, but they differ significantly from DOD.
Phrase Detection~\cite{plummer2020phrasedet} lacks explicit negative labels as negative instances are unlabeled, and does not constitute a true detection task. Additionally, its references are simply phrases.
In contrast, DOD ensures exhaustive annotation of positive and negative labels, and its references can be words, phrases, or sentences.
Cops-Ref benchmark~\cite{chen2020copsref} focuses on evaluating the grounding capability of REC methods in difficult negative regions with related and distracting targets. It provides explicit negative certificates for only a limited set of images.
In \ddd{}, negative certificates are available across the entire dataset.
Zero-shot grounding~\cite{sadhu2019zeroshotgrounding} centers on locating concepts not in the training set. It assumes the existence of the object referred by a reference in a image, and locates a single target per image, with a short phrase, while DOD makes no assumptions about the existence of the target, and locates zero to multiple targets, with varied expressions.

\subsection{Current methods}

\noindent \textbf{Open-vocabulary object detection methods.}
% Traditional object detection methods.
% Open vocabulary detection.
% Models integrating other similar tasks: 
% RegionCLIP~\cite{zhong2022regionclip}: adopt the pretrained visual encoder in CLIP and finetune on image-text pairs in CC3M.
% Detic. use clip embeddings to encode category names for OVD.
% GLIP and GLIP v2: phrase localization on flick30k.
% Object detection is a computer vision task that allows to identify and locate objects in an image or video.
% Object detection models have been traditionally formulated for closed-vocabulary settings, which can be divided as ``one-stage”~\cite{} and
% ``two-stage”~\cite{} formulations. 
% However, this vanilla object detection is limited to the fixed object category defined in advance, which greatly limits the practicability of existing methods.
Open-vocabulary detection is currently receiving increased attention. It aims to detect arbitrary classes using language for generalization, even when trained on a limited set of classes.
The first approach, OVR-CNN~\cite{zareian2021open}, utilizes image-caption pairs for pretraining the visual encoder to enhance its zero-shot generalization capabilities. With the introduction of CLIP~\cite{radford2021learning}, models such as Detic~\cite{zhou2022detecting}, DetCLIP~\cite{yao2022detclip}, RegionCLIP~\cite{zhong2022regionclip}, and OV-DETR~\cite{zang2022open} have further advanced image and language embeddings pretrained using CLIP.
ViLD~\cite{gu2022openvocabulary} further distills knowledge from CLIP to inherit language semantics for recognizing novel classes.
GLIP~\cite{li2022grounded,zhang2022glipv2} formulates object detection as a phrase grounding problem~\cite{plummer2015flickr30k} and utilizes additional phrase grounding data to facilitate vision-language alignment.

% It shows that such a formulation can even achieve stronger performance on fully-supervised detection benchmarks.

\noindent \textbf{Referring expression comprehension methods.}
% REC is natively a vision-language task.
Existing works~\cite{deng2021transvg,song2021co,li2021referring,liu2023polyformer,wang2022ofa} can be divided into three categories.
(1) Specialist models tailored for REC.
Previously, two-staged works~\cite{hu2017modeling,yu2016modeling} reformulate this as a ranking task.
More recently, one-stage approaches~\cite{zhou2021real,subramanian2022reclip,arbelle2021detector} speed up the inference process.
(2) Multi-task models~\cite{zhu2022seqtr,li2021referring,liu2023polyformer}.
They usually design a unified formulation for a few closely related tasks.
For example, SeqTR~\cite{zhu2022seqtr} unifies REC and RES as a point prediction problem.
% RefTR~\cite{li2021referring} proposes a simple one-stage multi-task framework for REC and RES, which is capable of simultaneous language grounding at both a bounding box and segmentation level.
% MCN~\cite{luo2020multi} novelly proposes Consistency Energy Maximization to enable REC and RES to focus on similar visual regions by maximizing the consistency energy between these two tasks.
% PolyFormer~\cite{liu2023polyformer} formulates REC and RES as a sequence-to-sequence prediction problem, which naturally fuses multi-modal features together as input and generate a sequence of polygon vertices and bounding box corner points.
(3) Multi-modal pre-training models~\cite{chen2020uniter,lu2022unified,wang2022ofa}.
% Recent progress in multi-modal understanding has been mainly powered by pre-training large transformer models to learn generic multi-modal representations from enormous amounts of aligned image-text data, then fine-tuning them on downstream tasks. 
% A prevalent paradigm is to extract visual and textual features independently and then use the attention mechanism of the transformers to learn an alignment between the two.
Unified-IO~\cite{lu2022unified} and OFA~\cite{wang2022ofa} propose unified sequence-to-sequence frameworks that can handle a variety of vision, language, and multi-modal tasks. Currently, OFA holds the SOTA among REC methods.

% These methods can be
% divided into single stream [6, 24, 65, 22] and two-stream
% [47, 28, 29, 46] architectures depending on whether the text
% and images are processed by a single combined transformer
% or two separate transformers followed by some cross attention layers. For both these types, the prevalent approach is
% to extract visual and textual features independently and then
% use the attention mechanism of the transformers to learn an
% alignment between the two

% Specialist models.
% TransVG.
% RefTR. SeqTR. PolyFormer. REC + RES.
% Pretrained models.
% OFA. UNITER. Unified-IO. multi-task pretrain + single-task fine-tuning.

\noindent \textbf{Bi-functional models for REC and OVD/OD.}
% Pretrain + Finetune.
% M-DETR. pretrain on detection data. finetune and eval on RefCOCO.
% Grounding-DINO. UNINEXT. train on multi-task data including detection and REC. direct evaluation on each task.
% \Todo{explain why these methods have not solve this problem.}
% OVD (or OD) and REC can be unified as instance perception tasks, which aims at finding certain objects specified by some queries such as category names, language expressions, and target annotations. Therefore, a lot of recent work attempts to combine OD and REC tasks, considering the instance perception tasks as ``text-conditioned object detection".
% Specifically, the OD task is used to pre-train a powerful object detector, and then the language modality is introduced to fine-tune a vision-language understanding specialized model (e.g. REC task). 
Some recent works~\cite{kamath2021mdetr,dou2022fiber,kuo2022findit,liu2023groundingdino,yan2023universal} aim to handle tasks such as OVD (or OD) and REC concurrently within a single model. They typically restructure the training approach for these tasks, enabling a single model to learn from datasets related to both tasks. However, the inference process for each task remains distinct and independent of the other.
% MDETR~\cite{kamath2021mdetr}, which is derived from the DETR~\cite{carion2020end} framework, undergoes pretraining on OD and image-text tasks, followed by fine-tuning and evaluation on REC.
FIBER~\cite{dou2022fiber} employs a two-stage pretraining strategy, separately utilizing image-text and image-text-box data to enhance data efficiency.
More recently, Grounding-DINO~\cite{liu2023groundingdino} extends a closed-set detector by performing vision-language fusion at multiple stages and evaluating its performance on REC datasets.
UNINEXT~\cite{yan2023universal} reformulates various image and video tasks into a unified object discovery and retrieval paradigm.
Despite these models sharing knowledge between detection and REC through pretraining, they are still treated as distinct tasks in these bi-functional models.

Methods with potential for DOD are continuously emerging and we will update them in \href{https://github.com/Charles-Xie/awesome-described-object-detection}{this list}.
% The main challenge in this task is how to transfer the image-level representations of the
% image-text backbone to detection despite the scarcity of localized annotations for
% rare classes. Making efficient use of the image-text pre-training is crucial since
% it allows for scaling without the need for expensive human annotations.
% The key solution of openset object detection is introducing language to a closed-set
% detector for open-set concept generalization.
% but this complete field has been split into multiple independent subtasks.

\section{Dataset}
\label{sec:dataset}
\vspace{-5pt}
\subsection{Dataset highlight}

\begin{figure}[t]
% \begin{center}
%    % \includegraphics[width=\linewidth]{figures/highlight.pdf}
%     \begin{subfigure}[b]{0.45\textwidth}
%     \centering
%         \includegraphics[width=\textwidth]{figures/highlight/highlight1.pdf}
%         \caption{Complete annotation.}
%         \label{fig:highlight1}
%     \end{subfigure}
%     \hfill
%     \begin{subfigure}[b]{0.25\textwidth}
%     \centering
%         \includegraphics[width=\textwidth]{figures/highlight/highlight2.pdf}
%         \caption{Unrestricted reference.}
%         \label{fig:highlight2}
%     \end{subfigure}
%     \hfill
%     \begin{subfigure}[b]{0.30\textwidth}
%     \centering
%         \includegraphics[width=\textwidth]{figures/highlight/highlight3.pdf}
%         \caption{Absence description.}
%         \label{fig:highlight3}
%     \end{subfigure}
%     \hfill
% \end{center}

% preliminary
\sbox\twosubbox{%
  \resizebox{\dimexpr1.0\textwidth-1em}{!}{%
    \includegraphics[height=3cm]{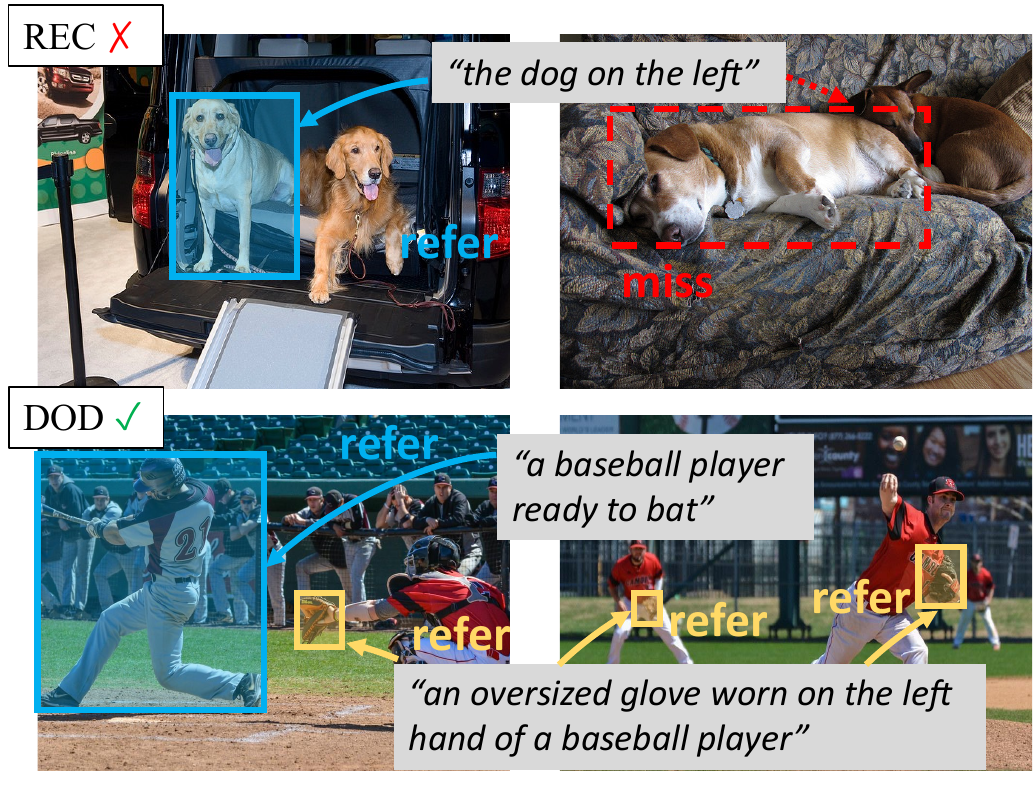}
    \includegraphics[height=3cm]{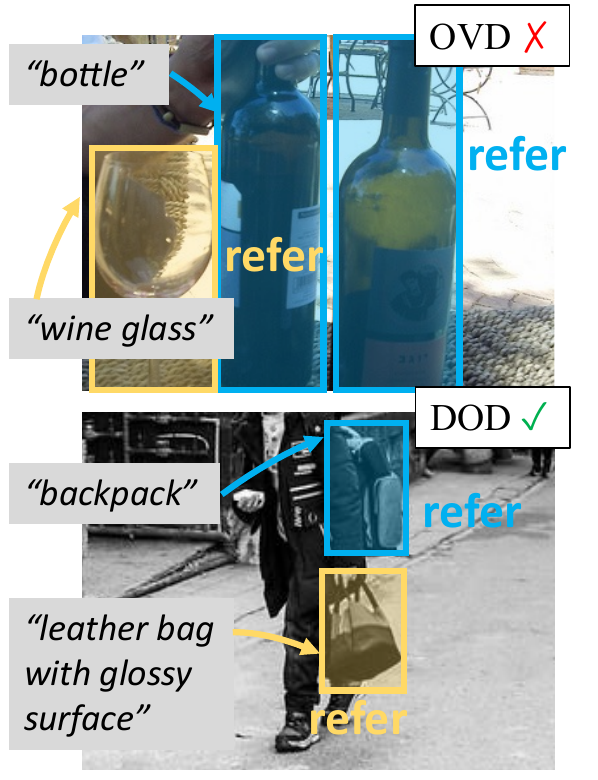}
    \includegraphics[height=3cm]{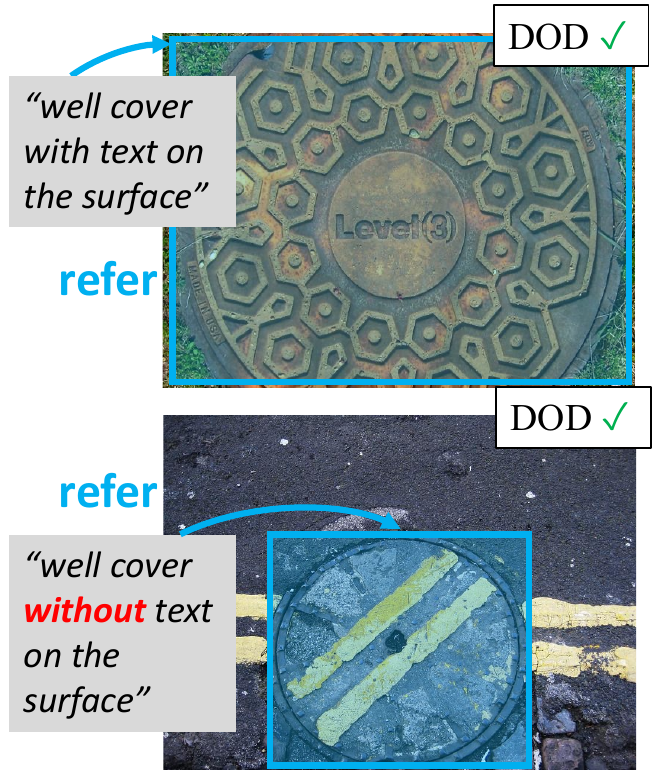}
  }%
}
\setlength{\twosubht}{\ht\twosubbox}

% typeset
\centering
\subcaptionbox{Complete annotation.\label{fig:highlight1}}{%
  \includegraphics[height=\twosubht]{figures/highlight/highlight1.pdf}%
}\quad
\subcaptionbox{Unrestricted reference.\label{fig:highlight2}}{%
  \includegraphics[height=\twosubht]{figures/highlight/highlight2.pdf}%
}
\subcaptionbox{Absence expression.\label{fig:highlight3}}{%
  \includegraphics[height=\twosubht]{figures/highlight/highlight3.pdf}%
}
% \vspace{-0pt}
\caption{Some examples from previous datasets and the proposed \ddd{} dataset for DOD. (a) Our dataset for DOD is completely annotated for detection, while REC datasets like RefCOCO are not. (b) Our dataset has unrestricted reference, while OVD datasets like COCO are not. (c) Our dataset not only provides traditional presence descriptions, but also absence descriptions.}
% \vspace{-18pt}
\label{fig:highlight}
\end{figure}

The proposed dataset is re-annotated on GRD~\cite{wu2023gres}, a dataset for RES~\cite{zhu2022seqtr,liu2023polyformer}. As briefly introduced in \cref{sec:introduction}, it contains three major characteristics. In \cref{fig:highlight}, we show some examples from previous datasets and \ddd{} to highlight them. Here we elaborate on them with a few other characteristics:

The first is \textit{complete annotation}. For REC, the instances referred to by one description are only annotated in a few images. For other images without the annotation of this description, it is unknown whether the corresponding instance exists or not. That is to say, their annotations are not complete. Contrarily, as shown in \cref{fig:highlight1}, in \ddd{}, the objects referred to in all images by any description are annotated, as are the negative samples, like traditional object detection datasets.

\begin{figure}[t]
\begin{center}
    \begin{subfigure}[b]{0.24\textwidth}
     \centering
     \includegraphics[width=\textwidth]{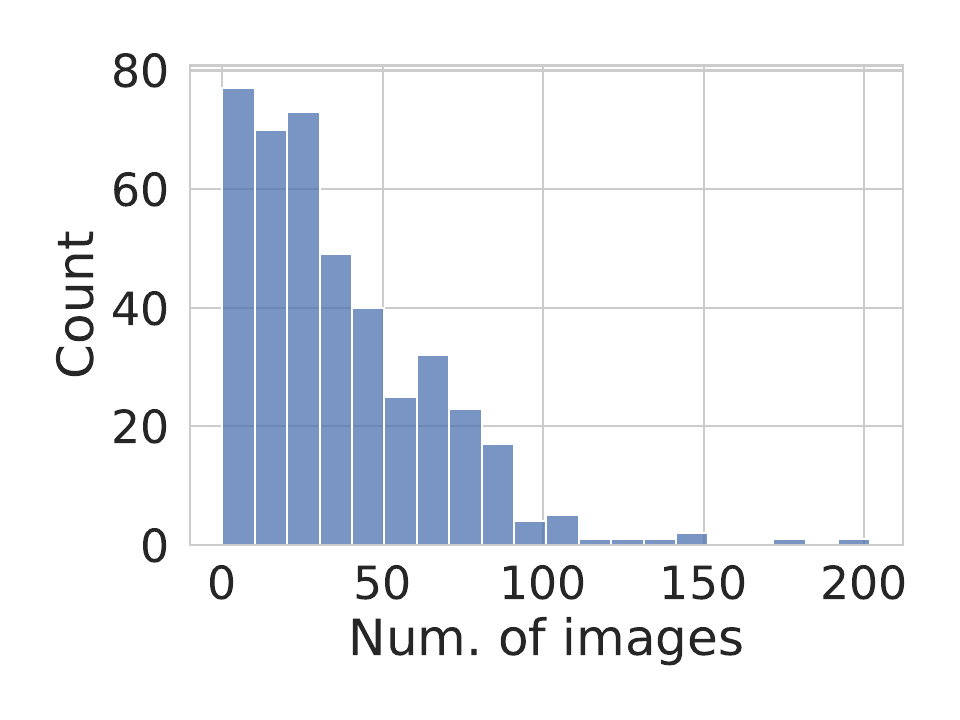}
     \caption{
     % Num. of positive images for a description.
     }
     \label{fig:nimg_per_ref}
    \end{subfigure}
    \hfill
    \begin{subfigure}[b]{0.24\textwidth}
     \centering
     \includegraphics[width=\textwidth]{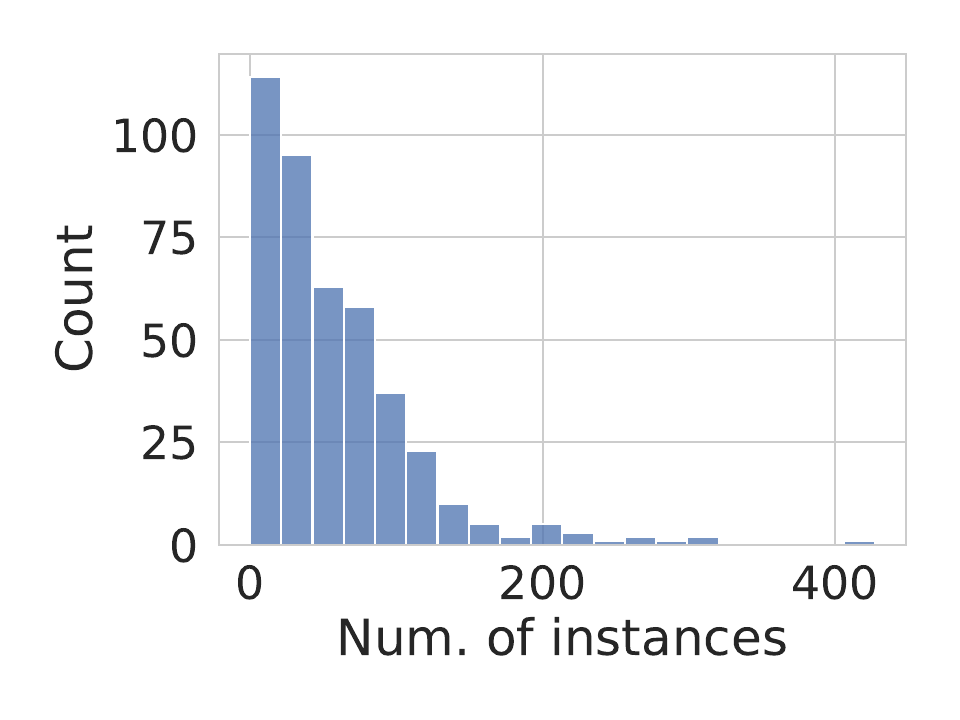}
     \caption{
     % Num. of positive instances for a description.
     }
     \label{fig:nbox_per_ref}
    \end{subfigure}
    \hfill
    \begin{subfigure}[b]{0.24\textwidth}
     \centering
     \includegraphics[width=\textwidth]{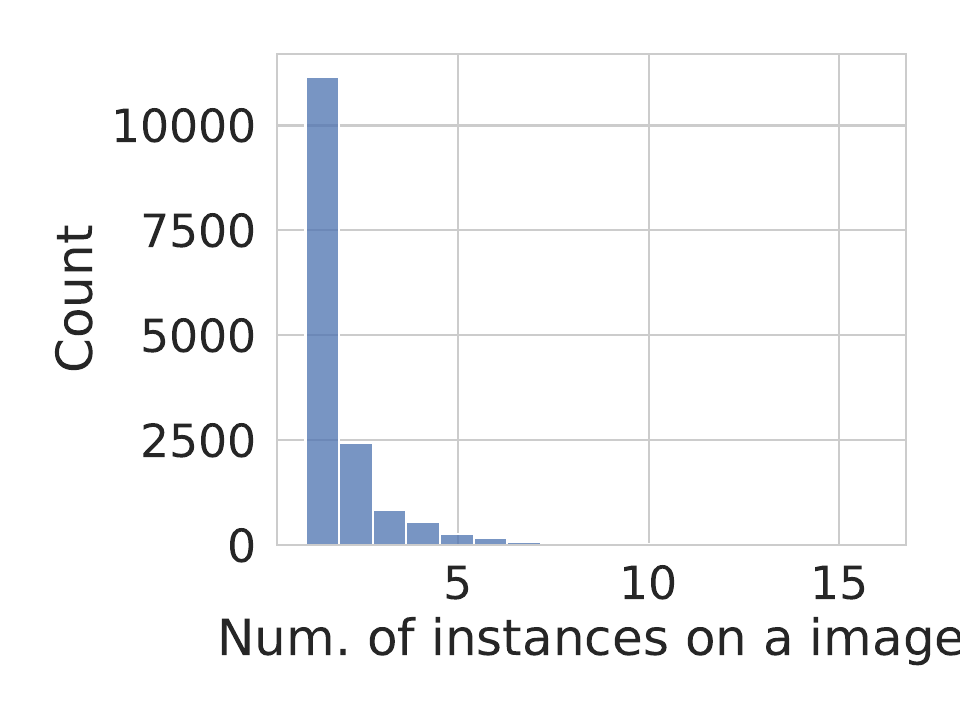}
     \caption{
     % Num. of instances in a positive image for a description.
     }
     \label{fig:nbox_per_img_ref}
    \end{subfigure}
    \hfill
    \begin{subfigure}[b]{0.24\textwidth}
     \centering
     \includegraphics[width=\textwidth]{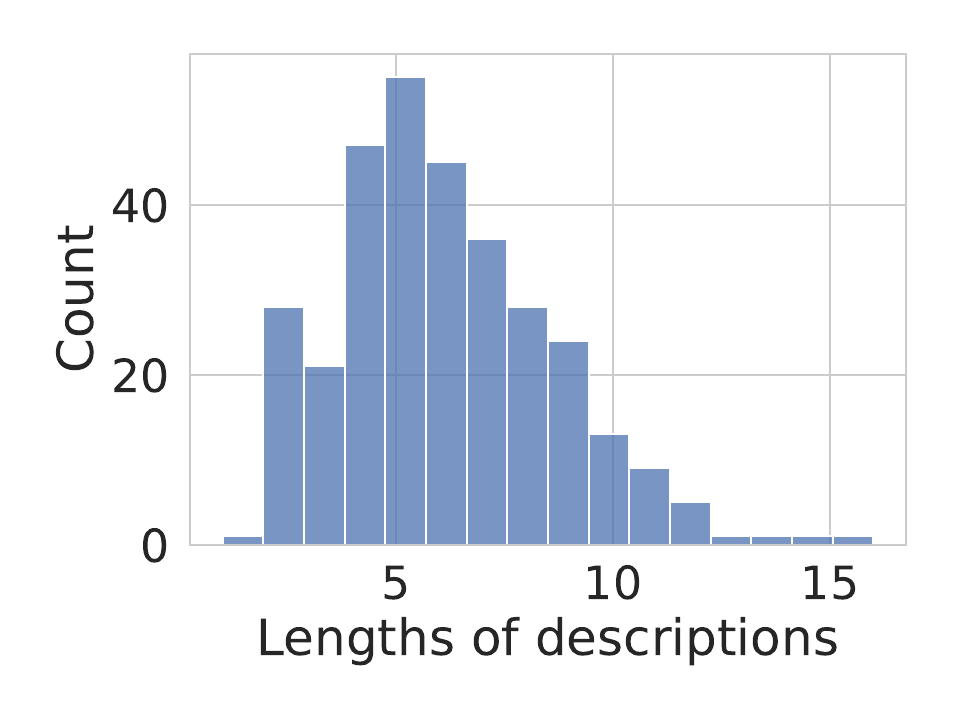}
     \caption{
     % Lengths of descriptions in the dataset.
     }
     \label{fig:len_refs}
    \end{subfigure}
\end{center}
   \caption{Distribution of (a) number of positive images for a description in the dataset, (b) number of positive instances for a description, (c) number of instances in a positive image for a description, and (d) lengths of descriptions.}
\label{fig:lengths_nums}
\end{figure}

The second is \textit{unrestricted language description}. As shown in \cref{fig:highlight2}, unlike (open vocabulary) object detection that retrieves objects with category names, we retrieve objects with language expressions, which is rather flexible. As is shown in \cref{fig:len_refs}, the lengths of descriptions in \ddd{} vary a lot. The shortest descriptions have one or two words, where the DOD task downgrades to OVD, while the longest may have 15 or more words, resulting in rather complex language expressions.

The third is \textit{absence expression}. Current datasets with language description, like RefCOCO series for REC, usually describe objects with certain features. They usually focus on the ability to discover the existence of concepts but neglect their absence. Noticing the missing ability to verify such capability, we also annotate objects lacking a certain attribute. \cref{fig:highlight3} shows an example with presence description and another with absence description from \ddd{}. Such absence description makes up about one quarter of the references in this dataset. This is a first for existing benchmarks.

The fourth is \textit{instance-level annotation}, a characteristic not held by GRD as it is intended for RES.

The fifth is \textit{one description can refer to multiple instances} in an image, as in \cref{fig:nbox_per_img_ref}. This is not true for REC datasets. If we regard category names as references, then OD datasets do have this feature.

In summary, the proposed dataset differs from the REC dataset primarily in terms of characteristics 1st, 3rd, and 5th. In contrast, when compared to OD datasets, the proposed dataset showcases disparities in the 2nd and 3rd characteristics, and when compared with GRD, in the 2nd, 3rd, 4th, and 5th characteristics. We refer the readers to the \supp{} for more information about the characteristics of \ddd{} and more examples.

\subsection{Annotation process}

We utilize the GRD dataset~\cite{wu2023gres} as the source for images, along with its original annotations. Originally, it is divided into multiple groups, each containing several references, with positive and negative samples annotated only within each group. We extend the annotations in three aspects:

\noindent \textbf{Adding instance-level annotation.}
GRD is designed for RES, where each reference corresponds to one semantic mask across one image. However, for the DOD task, which requires the recognition and localization of individual instances, we annotate each instance referred to by a description with an individual bounding box (along with an instance mask). This is the basic step to adapt the dataset for instance localization.

\noindent \textbf{Adding complete annotations.}
In addition to the intra-group annotation in GRD, we further annotate the positive and negative samples for each reference across the entire dataset. With complete dataset-wise annotations, the division into groups becomes unnecessary for evaluation, serving only as a means to organize references by scenarios. This enhancement makes the dataset suitable for detection tasks, significantly increasing the number of positive and negative samples.

Note that we use the complete annotation similar to COCO~\cite{lin2014microsoft}, i.e., explicit positive and negative certificates for all categories on all images, rather than federated annotation~\cite{gupta2019lvis,Krasin2017}. This allows using mAP (mean Average Precision) as the evaluation metric, which is elaborated in \cref{sec:eval_metric}.

\noindent \textbf{Adding annotations for absence expressions.}
We have designed many absence descriptions based on the scenarios within the dataset, in addition to the traditional presence expressions in GRD. We annotate the instances in the images across the entire dataset with these absence expressions.
This step increases the difficulty level of the proposed benchmark and enables the evaluation of existing models' ability to comprehend the absence of concepts.

We present a concise overview of the overall annotation process here.
We organize groups of images and references (both for presence and absence). For each image, the references in its group are used. References from other groups may also appear, but with lower probability.
We employ CLIP~\cite{radford2021learning} to select a large number of candidates from these references in other groups.
We manually check and adjust the hyper-parameters to make sure that such CLIP filtering usually do not miss positive refs.
Subsequently, annotators select the positive references from these candidates (rather than from all references in the dataset) and add bounding boxes to the images.
For more detailed information regarding the annotation process, please refer to \supp{}.

\subsection{Dataset statistics}

\noindent \textbf{GRD statistics.}
% {'nsent': 316, 'nanno': 18343, 'nimg': 10634, 'ngroup': 106, 'num_img_sent': 9238, 'num_anti_img_sent': 22454, 'num_anno_sent': 13734, 'num_anti_anno_sent': 40901, 'avg_sent_len': 5.550632911392405}
% {'nsent': 422, 'nanno': 18343, 'nimg': 10634, 'ngroup': 106, 'num_img_sent': 9238, 'num_anti_img_sent': 22454, 'num_anno_sent': 13734, 'num_anti_anno_sent': 40901, 'avg_sent_len': 5.890995260663507}
It has 10,578 images collected online, divided into 106 groups.
Each group has around 100 images and 3 expressions referring to segmentation masks in this group, resulting in 316 references, 9,323 positive image-text pairs and 22,201 negative pairs.
Note that it only annotates positive and negative samples inside each group, i.e., the annotation completeness is only \textbf{group-level}, so a reference will not be verified outside its group.
The expressions have an average length of 5.9 words.
We refer the reader to the original paper for specific statistics of GRD.

\noindent \textbf{\ddd{} statistics.}
The proposed \ddd{} has 10,578 images, all from GRD. It has 422 well-designed expressions, including 316 expressions from GRD and 106 absence expressions we added (one for each scenario).
The instance-level annotation results in 18,514 boxes.

% intra-group:
% {'nsent': 316, 'nanno': 18514, 'nimg': 10578, 'ngroup': 106, 'num_img_sent': 9323, 'num_anti_img_sent': 22201, 'num_anno_sent': 13917, 'num_anti_anno_sent': 41231, 'avg_sent_len': 5.987341772151899}
% {'nsent': 422, 'nanno': 18514, 'nimg': 10578, 'ngroup': 106, 'num_img_sent': 13035, 'num_anti_img_sent': 29067, 'num_anno_sent': 20279, 'num_anti_anno_sent': 53383, 'avg_sent_len': 6.31042654028436}
% inter-group:
% {'nsent': 316, 'nanno': 18514, 'nimg': 10578, 'ngroup': 106, 'num_img_sent': 9323, 'num_anti_img_sent': 22201, 'num_anno_sent': 16480, 'num_anti_anno_sent': 5833944, 'avg_sent_len': 5.987341772151899}
% {'nsent': 422, 'nanno': 18514, 'nimg': 10578, 'ngroup': 106, 'num_img_sent': 13035, 'num_anti_img_sent': 29067, 'num_anno_sent': 24282, 'num_anti_anno_sent': 7788626, 'avg_sent_len': 6.31042654028436}
Due to the effort in \textit{complete annotation}, for a reference, each image in the dataset is annotated for possible positive and negative samples, i.e., the annotation completeness is \textbf{dataset-level}.
Thus, there are 24,282 positive object-text pairs and 7,788,626 negative pairs, orders of magnitude larger than GRD.
Among them, those with images and texts from the same scenario are probably more difficult, which includes 20,279 positive and 53,383 negative pairs.
The average length of expressions is 6.3 words, due to the relative longer absence expressions.
More statistics and examples of \ddd{} are available in \supp{}.

\subsection{Evaluation metrics}
\label{sec:eval_metric}

The classification of instances in \ddd{} is \textbf{multi-label}. Each description corresponds to a category. Naturally, there can be relationships between categories, such as parent-child hierarchies, synonyms, and partial overlap. When designing categories, we intentionally reduce parent-child or synonym relationships to ensure greater diversity and challenge. However, there exists partial overlap between categories. Therefore, in \ddd{}, one instance may correspond to multiple descriptions, and the classification in \ddd{} is multi-label~\cite{gupta2019lvis} rather than single-label~\cite{lin2014microsoft}, making it suitable for categories with relationships. An effective detector should assign all relevant positive categories (e.g., \texttt{dog not led by rope outside} and \texttt{clothed dog} for a clothed dog not led by rope outside) for an instance.

We use \textbf{standard mAP} for evaluation. Given the multi-label setting and the exhaustive annotation (all positive and negative labels are known for an instance) of \ddd{}, category relationships will not affect the evaluation, so we can use consistent evaluation for each category across all images.
We describe the evaluation process here.
For inference, an instance predicted with category A and B is regarded as an instance for category A and an instance for B.
The AP for each category is computed as follows: Predictions for each category across all images are sorted by score in descending order, and those with a ground truth IoU exceeding a threshold are counted as TP (and the ground truth is marked as taken), while the rest are counted as false positives. With these TP and FP instances, we calculate the precision, recall, and AP. The mAP is calculated by averaging the AP across all categories.

We use \textsl{FULL}, \textsl{PRES}, and \textsl{ABS} to denote evaluation on all descriptions, presence descriptions only, and absence descriptions only. If not noted explicitly, the \textsl{FULL} setting is adopted.
The specific metrics for \ddd{} include:
\textit{Intra-scenario mAP:}
For this metric, we perform evaluation on each image with only the descriptions from the image's scenario. The final metric is the mAP averaged on different IoU thresholds from 0.5 to 0.95, following COCO~\cite{lin2014microsoft}. This is used as the default metric in our experimental settings.
\textit{Inter-scenario mAP:}
It is similar to the intra-scenario mAP described above, except that for each image, we detect the possible instances with all 422 references. This is aligned with the common mAP in object detection datasets~\cite{lin2014microsoft} and is much more challenging than the intra-scenario mAP.

% \noindent \textbf{Recall.}
% In REC datasets like RefCOCO~\cite{yu2016modeling,mao2016generation}, each image will have one and only one prediction for each reference, and the standard metric is accuracy (which is precision and also recall). This is not suitable for DOD, which is essentially a detection task. We adopt the average recall metric in COCO API for some analyses, but it does not necessarily correspond to the effectiveness of a method.

% The evaluation process for this metrics are illustrated in \cref{fig:evaluation}.

\section{Baselines}
\label{sec:baselines}

\subsection{Existing baselines from different tasks}
\label{subsec:existing_baselines}

We choose multiple advanced methods to verify on \ddd{} from OVD, REC to bi-functional methods. More details of these methods and their inference process are in our \supp{}.

\noindent \textbf{REC methods.}
We employ the state-of-the-art REC method, OFA~\cite{wang2022ofa}, with two variants.
OFA is based on an encoder-decoder, sequence-to-sequence framework. It is a multi-modal multi-task generalist that deals with different tasks together and is trained on various tasks, including language tasks (masked language modeling), image-to-text tasks (image captioning and Visual Question Answering (VQA)), and localization tasks (REC). Notably, although it is trained with a detection dataset, it is not evaluated on object detection and achieves poor performance if we do. Currently, it holds the SOTA performance on standard REC benchmarks like the RefCOCO series.

\noindent \textbf{OVD methods.}
We evaluate OWL-ViT~\cite{minderer2022simpleOWLViT} with two variants and CORA~\cite{wu2023cora}. They are the SOTA methods on OVD tasks, with a vision transformer as well as a language transformer. They are pretrained with image-text contrastive learning and then fine-tuned on detection dataset.

\noindent \textbf{Bi-functional methods utilizing both REC and OVD data.}
Methods falling into this category are not many but emerging fast recently. We test two methods: Grounding-DINO~\cite{liu2023groundingdino} and UNINEXT~\cite{yan2023universal}, each with two variants. Both of them are based on DETR~\cite{carion2020end}. They are pretrained on multiple datasets, including detection and REC datasets, and then evaluated with different strategies for different tasks.

\subsection{A proposed baseline}
\label{subsec:our_baseline}

\ddd{} is very challenging for existing works, as we will demonstrate in \cref{subsec:experimental_comp}. We have selected one of these works for adjustment to provide a better baseline.
The chosen work should (1) be capable of understanding text of various lengths; (2) excel in their original tasks; (3) have a framework with a rather simple technical design, allowing us to modify its components easily. We have chosen OFA because it (1) is a multi-modal multi-task framework with MLM (Masked Language Modeling) and image-to-text pretraining; (2) achieves SOTA on REC; (3) has a simple seq2seq framework.

However, OFA faces several problems that make it unsatisfactory for this task, as discussed in \cref{sec:experiments}. First, forcing multiple tasks of different modalities into one seq2seq framework adversely affects the performance of specific tasks, especially tasks related to localization. Second, training on the grounding task results in poor ability to handle multiple instances. We evaluated the model on COCO detection, and it achieved less than 10 mAP. Thirdly, its REC paradigm also makes it predict only one instance, making it unable to reject negative images and irrelevant descriptions.

Therefore, we have made some modifications to OFA to make it more suitable for this task.
The first modification is \textbf{granularity decomposition} to make it more suitable for localization. We have divided the pretraining tasks of OFA into two different granularities: global tasks (related to language modeling, such as captioning, VQA, MLM, etc.) and local tasks (related to localization, such as detection and REC). We have added an additional decoder parallel to the original decoder in OFA that handles the local tasks, while the original decoder focuses on the global tasks. This alleviates conflicts between different tasks and enhances localization.

The second modification is \textbf{reconstructed data} for pretraining on REC, aiming to improve multi-target localization. We have reconstructed the data for REC to ensure that (1) multiple references are input for an image, and (2) a reference does not necessarily correspond to one object, but zero or multiple. This results in a unified data format for detection and REC, although the labels may be noisy since they were not initially prepared for DOD.

The third modification is \textbf{task decomposition} to empower the model with the ability to reject false positives. We have reformulated the training on reconstructed data into two tasks: REC (for locating a region based on a reference) and VQA (for determining if a region and a reference match each other, essentially a binary classification). The second step is responsible for rejecting false positives.
% For inference, we perform REC and VQA sequentially for each reference.

We refer to the model with all three modifications as \textbf{OFA-DOD}. More details on the proposed improvements can be found in the \supp{}.
It is important to note that this model is far from perfect for the complex \ddd{} benchmark. As we will show in \cref{subsec:experimental_comp}, although it outperforms existing methods, it serves as a baseline for future tasks on \ddd{}.

\section{Experimental Analyses}
\label{sec:experiments}

\subsection{Comparison of baselines on our metrics}
\label{subsec:experimental_comp}

\begin{table}[t]
    \centering
    \caption{Comparison of different methods on the proposed dataset for different mAP metrics.}
    \begin{tabular}{l l | c c c | c c c}
        \hline
        \multirow{2}{*}{Task} & \multirow{2}{*}{Method} & \multicolumn{3}{c}{Intra-scenario} & \multicolumn{3}{| c}{Inter-scenario} \\
        & & \textsl{FULL} & \textsl{PRES} & \textsl{ABS} & \textsl{FULL} & \textsl{PRES} & \textsl{ABS} \\
        \hline
        \multirow{2}{*}{REC} & OFA$_\text{base}$ & 3.4 & 3.0 & 4.3 & 0.1 & 0.1 & 0.1 \\
        & OFA$_\text{large}$ & 4.2 & 4.1 & 4.6 & 0.1 & 0.1 & 0.1 \\
        \hline
        \multirow{3}{*}{OVD} & CORA$_\text{R50}$ & 6.2 & 6.7 & 5.0 & 2.0 & 2.2 & 1.3 \\
        & OWL-ViT$_\text{base}$ & 8.6 & 8.5 & 8.8 & 3.2 & 3.7 & \textbf{4.7} \\
        & OWL-ViT$_\text{large}$ & 9.6 & 10.7 & 6.4 & 2.5 & 2.9 & 2.1 \\
        \hline
        \multirow{4}{*}{Bi-functional} & UNINEXT$_\text{large}$ & 17.9 & 18.6 & 15.9 & 2.9 & 3.1 & 2.5 \\
        & UNINEXT$_\text{huge}$ & 20.0 & 20.6 & 18.1 & 3.3 & 3.9 & 1.6 \\
        & G-DINO$_\text{tiny}$ & 19.2 & 18.5 & 21.2 & 2.3 & 2.5 & 2.1 \\
        & G-DINO$_\text{base}$ & 20.7 & 20.1 & \textbf{22.5} & 2.7 & 2.4 & 3.5 \\
        \hline
        DOD & OFA-DOD$_\text{base}$ & \textbf{21.6} & \textbf{23.7} & 15.4 & \textbf{5.7} & \textbf{6.9} & 2.3 \\
        \hline
    \end{tabular}
    \label{tab:comparison}
\end{table}

We make comparisons on the baselines introduced in \cref{sec:baselines}, mainly with the intra-scenario setting.
Unless explicitly noted, this is the default setting, instead of the more difficult inter-scenario.

\noindent \textbf{Existing SOTAs are insufficient for DOD, and bi-functional models outperform others.}
As demonstrated in \cref{tab:comparison}, existing methods, while achieving SOTA performance on their original benchmarks, fall short in delivering strong performance on \ddd{}.
Among them, recent bi-functional methods~\cite{liu2023groundingdino,yan2023universal} are notably superior to others, and currently, OVD methods outperform REC.
The inferiority of REC methods is likely due to their impractical setting described in \cref{sec:introduction}, which involves predicting one and only one instance for each reference. We will delve into this further.

\noindent \textbf{Rejecting irrelevant references are difficult, which REC are naturally incapable of.}
In contrast to intra-scenario evaluation, the inter-scenario setting assesses all references in the dataset for each image. Since references from other scenarios are likely not semantically relevant to the images, this necessitates the ability to reject irrelevant references for an image. This aligns with the evaluation in standard detection tasks.
From \cref{tab:comparison}, it is evident that OFA, a REC method, almost completely fails in this setting. This is caused by its prediction of a region for every reference, resulting in a large number of false positives when there are numerous candidate references. This underscores the importance of empowering REC methods with the ability to reject false positives.
We find that none of the verified methods achieve good performance under the inter-scenario setting, indicating that existing methods are far from being capable of DOD. This highlights the challenge of \ddd{}

\noindent \textbf{The proposed baseline outperforms existing methods.}
The proposed baseline is based on OFA, but our improvements significantly enhance its performance. It outperforms all existing methods in the intra-scenario setting and surpasses them by a wider margin in the inter-scenario setting. This may suggest that the proposed baseline has a stronger ability to reject irrelevant references.
Nonetheless, the proposed method is far from perfect and can only serve as a baseline for future research.

\begin{figure}[t]
\begin{center}
% \fbox{\rule{0pt}{2in} \rule{0.9\linewidth}{0pt}}
     \begin{subfigure}[b]{0.24\textwidth}
         \centering
         \includegraphics[width=\textwidth]{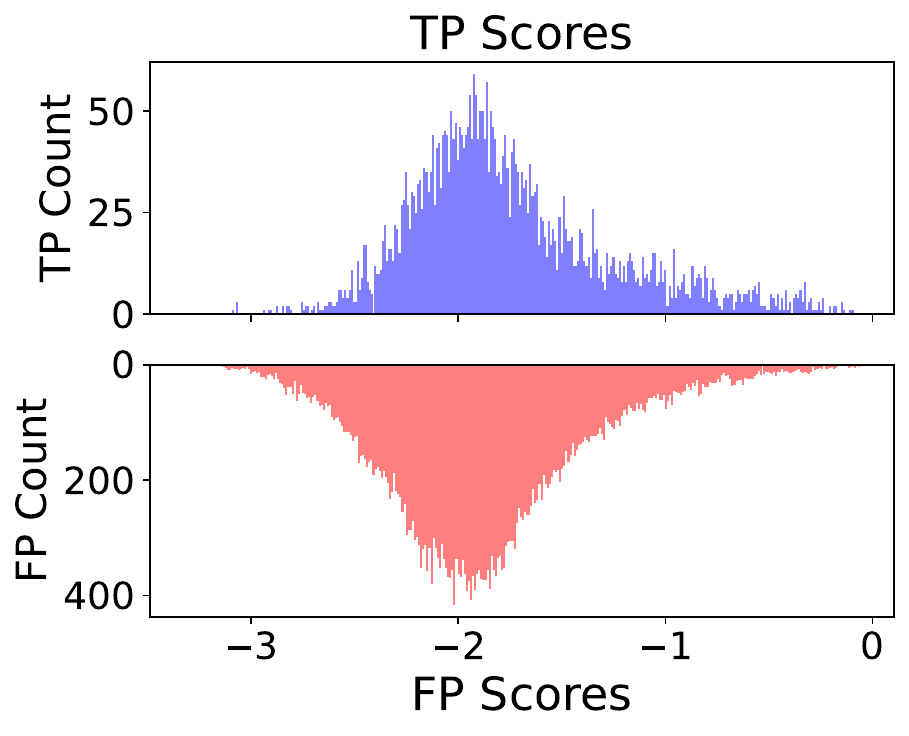}
         \caption{OFA}
         \label{fig:ofa_scores}
     \end{subfigure}
     \hfill
     \begin{subfigure}[b]{0.24\textwidth}
         \centering
         \includegraphics[width=\textwidth]{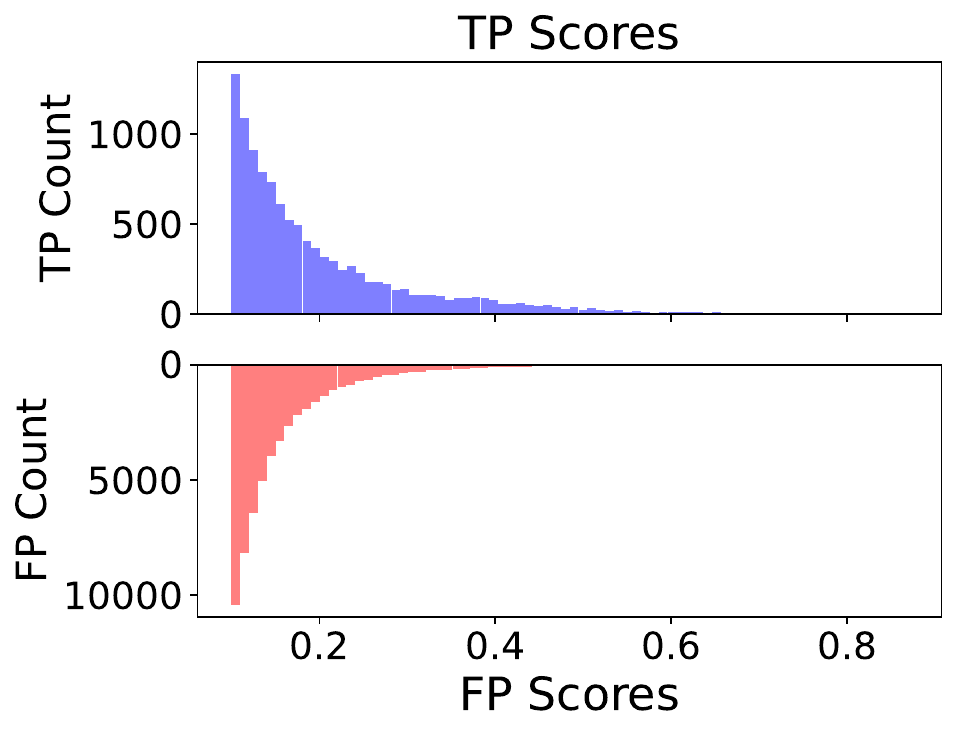}
         \caption{OWL-ViT}
         \label{fig:owlvit_scores}
     \end{subfigure}
     \hfill
     \begin{subfigure}[b]{0.24\textwidth}
         \centering
         \includegraphics[width=\textwidth]{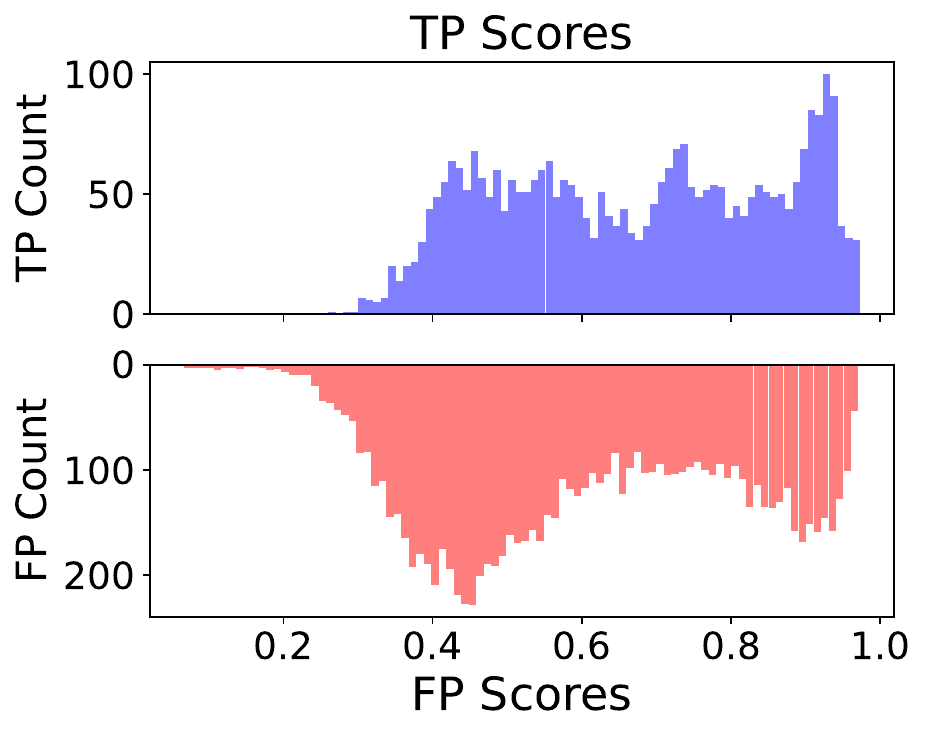}
         \caption{Grounding-DINO}
         \label{fig:gdino_scores}
     \end{subfigure}
     \hfill
     \begin{subfigure}[b]{0.24\textwidth}
         \centering
         \includegraphics[width=\textwidth]{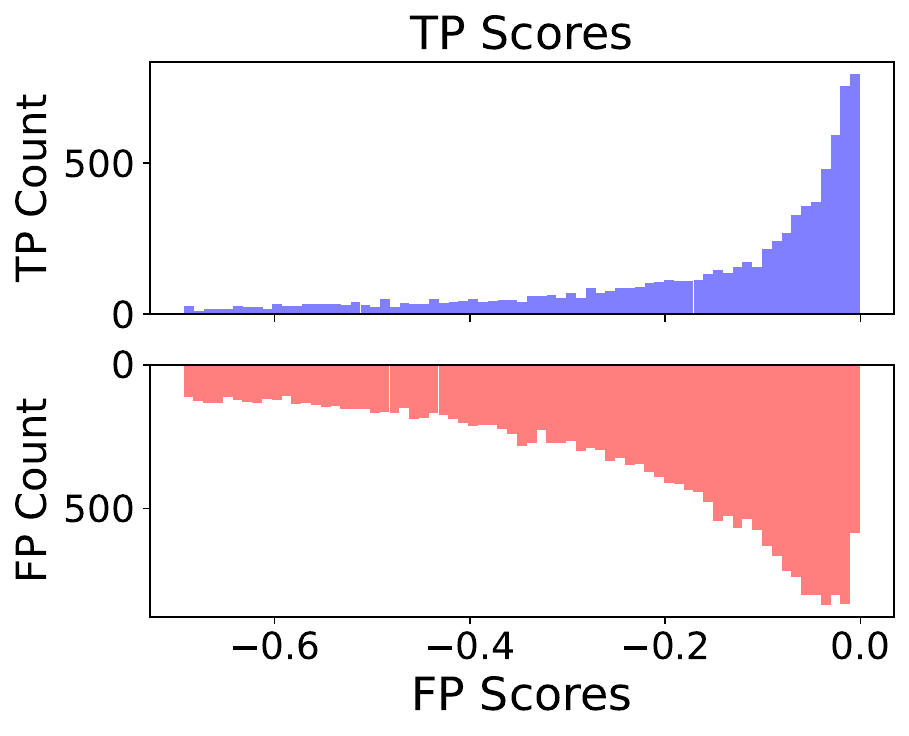}
         \caption{OFA-DOD}
         \label{fig:ofa_ours_scores}
     \end{subfigure}
\end{center}
   \caption{Distribution of TP and FP scores from different baseline methods.}
   % \vspace{-5pt}
\label{fig:score_distribution}
\end{figure}

\subsection{Further analysis}

\noindent \textbf{Absence descriptions are more difficult for most methods.}
As shown in \cref{tab:comparison}, the performance of baseline methods on \textsl{PRES} (presence descriptions) is consistently superior to that on \textsl{ABS} (absence descriptions). This suggests that existing methods may not effectively differentiate between the presence and absence of attributes in a language description.

\noindent \textbf{REC methods fail to provide good confidence scores.}
We visualized the score distributions from baselines for TPs and FPs, to assess their capabilities in classification and confidence estimation.
As in \cref{fig:score_distribution}, the confidence scores from OFA do not exhibit a clear distinction between TP and FP cases. This can be attributed in part to the seq2seq framework in OFA, which does not directly yield confidence scores, and in part to the grounding formulation of REC, which identifies the image region most similar to the text description without distinguishing between positive and negative.

With a task decomposition step to enhance binary classification performance, our OFA-DOD demonstrates a significant disparity between TP and FP score distributions, yielding more reliable classification results. Note that this improvement does not necessitate modifications to the model framework or training datasets; rather, it is attributed to a more appropriate task formulation.

\begin{table}[t]
    \centering
    \caption{Evaluation regarding different number of instances in a image for each reference.}
    % \resizebox{\textwidth}{!}{
    \begin{tabular}{l | c | c | c c c c}
        \hline
        Method & No-instance & One-instance & \multicolumn{4}{c}{Multi-instance mAP(\%)} $\uparrow$ \\
        & FPPC (\%) $\downarrow$ & mAP (\%) $\uparrow$ & 2 & 3 & 4 & 4+ \\
        \hline
        OFA & 100.0 & 14.8 & 9.5 & 7.9 & 5.4 & 3.7 \\
        CORA & 17.3 & 9.7 & 8.4 & 9.5 & 9.0 & 8.5 \\
        OWL-ViT & 41.9 & 21.1 & 17.3 & 16.6 & 16.0 & 14.0 \\
        UNINEXT & 100.0 & 55.7 & 26.2 & 18.6 & 14.4 & 9.0 \\
        G-DINO & 100.0 & 63.7 & 28.3 & 19.7 & 15.9 & 10.1 \\
        OFA-DOD & 35.6 & 56.4 & 19.6 & 12.7 & 10.3 & 7.1 \\
        \hline
    \end{tabular}
    % }
    % \vspace{-10pt}
    \label{tab:analysis_box_num}
\end{table}

\noindent \textbf{Multi-instance detection is challenging for methods other than OVD.}
For each image, \ddd{} can have zero to multiple instances \textbf{for a single description}. To assess how current methods handle varying numbers of instances, we conducted evaluations under three different settings: \textbf{no-instance}, where for a reference, evaluations are limited to images without any referred instance; \textbf{one-instance}, for images with a single instance; and \textbf{multi-instance}, for images with multiple instances. As shown in \cref{tab:analysis_box_num}, OVD methods outperform others when multiple instances are referred by the description, although they may not be as competitive on the entire dataset or images with few instances. Notably, OWL-ViT maintains consistent performance even as the number of instances increases, which sets it apart from other methods. In contrast, REC and current bi-functional methods struggle in multi-instance scenarios. This highlights the strength of OVD methods in multi-target detection, while REC and current bi-functional approaches are less robust in such situations.

\noindent \textbf{REC and bi-functional methods lack the ability to reject negative instances.} In the \textbf{no-instance} column of \cref{tab:analysis_box_num}, we do not report mAP since there are no positive instances in GT for the corresponding reference, making AP inapplicable. Predictions on such images are FPs, so we measure the ratio of images where FPs are produced to the total number of no-instance images for a given reference, namely False Positives Per Category (FPPC). We report the average FPPC over all references.
We observe that most baselines are incapable of determining whether an image contains the referred target or not, yet they still produce predictions. This behavior is expected for REC methods. Bi-functional methods, trained and inferred with the REC task formulation, also exhibit this issue. Only the OVD method and our proposed baseline can effectively reject such negative image-text pairs.

\begin{table}[t]
    \centering
    \caption{Evaluation one references with various lengths.}
    % \resizebox{\textwidth}{!}{
    \begin{tabular}{l | c c c c}
        \hline
        Method & \textit{short} & \textit{middle} & \textit{long} & \textit{very long} \\
        \hline
        OFA & 4.9 & 5.4 & 3.0 & 2.1 \\
        % CORA & 3.3 & 5.0 & 5.8 & 4.4 \\
        OWL-ViT & 20.7 & 9.4 & 6.0 & 5.3 \\
        UNINEXT & 18.5 & 23.3 & 17.4 & 16.1 \\
        G-DINO & 22.6 & 22.5 & 18.9 & 16.5 \\
        OFA-DOD & 23.6 & 22.6 & 20.5 & 18.4 \\
        \hline
    \end{tabular}
    % }
        % \vspace{-10pt}
    \label{tab:analysis_ref_length}
\end{table}

\noindent \textbf{OVD methods suffer from long descriptions greatly while others do not.}
We partition the references according to their lengths and then evaluate on these partitions. The results are shown in \cref{tab:analysis_ref_length}, where \textit{short}, \textit{middle}, \textit{long} and \textit{very long} corresponding to references with 1\textasciitilde 3, 4\textasciitilde 6, 7\textasciitilde 9, and more than 9 words.
For \textit{short} descriptions, which is close to OVD setting, OVD and bi-functional methods obtain similar performance. However, as the length of references increases, the performance of OVD methods decrease fast, while REC and bi-functional methods suffer less from this. We can see that OVD methods are sensitive to long references, as expected, while other two types do not.

More experiments and additional \textbf{qualitative results} are available in \supp{}.

% \input{tables/recall}

% \noindent \textbf{Compared with REC, OVD methods tend to make predictions only when it's certain.}
% % \noindent \textbf{Analysis on recalls.}
% The recalls of different methods are shown in \cref{tab:recall}. OVD methods are obviously bad at recall, which means that it tends to produce a result when it is quite certain. Grounding-DINO, though performs not as good as the proposed baseline in terms of mAPs, obtains the best recall. This indicates that it tends to produce more detection results.

% \Todo{do we need to consider inference strategy (one ref at a time vs all ref at a time? No.}

% \input{figures/qualitative}

% \noindent \textbf{Qualitative analysis.}
% We visualize the results of different baseline methods on the proposed benchmark. This is shown in \cref{fig:qualitative}. We can see that the proposed method handles such images better.

\begin{table}[t]
    \centering
    \caption{Ablation on the proposed baseline for its improvement components and the training data.}
    \begin{subtable}[t]{0.45\textwidth}
        \centering
        \caption{Method components.}
        \begin{tabular}{c c c c | c}
        \hline
        OFA & GD & RD & TD & mAP(\%) \\
        \hline
        \cmark & \xmark & \xmark & \xmark & 3.4 \\
        \cmark & \cmark & \xmark & \xmark & 10.5 \\
        \cmark & \cmark & \cmark & \xmark & 17.2 \\
        \cmark & \cmark & \cmark & \cmark & 21.6 \\
        \hline
       \end{tabular}
       % \vspace{-10pt}
       \label{tab:ablation_components}
    \end{subtable}
    \hfill
    \begin{subtable}[t]{0.45\textwidth}
        \centering
        \caption{Training data.}
        \begin{tabular}{c c c c | c}
        \hline
        REC & OD & I2T & MLM & mAP(\%) \\
        \hline
        \cmark & \cmark & \cmark & \cmark & 21.6 \\
        \cmark & \xmark & \cmark & \cmark & 16.4  \\
        \cmark & \cmark & \xmark & \cmark & 14.2 \\
        \cmark & \cmark & \cmark & \xmark & 20.3 \\
        \hline
       \end{tabular}
       \label{tab:ablation_tasks}
    \end{subtable}
    % \vspace{-10pt}
    \label{tab:method_ablation}
\end{table}

\subsection{Ablation on the proposed baseline}

\noindent \textbf{Method components.}
In \cref{tab:method_ablation}, we perform ablation on the proposed improvements in our baseline, step-by-step from OFA to OFA-DOD, to see how they affect the performance.
Granularity decomposition (GD) makes the method more suitable for localization task.
It disentangle tasks of global or local granularity by handling them with 2 separated branch.
Reconstructed data (RD) uniforms REC and OD data into the same form, and prepares multi-instance samples with both short and long references.
Task decomposition (TD) is proposed to help rejecting FPs.
It breaks down the DOD task into a REC step followed by a VQA step.
All three of them improve the performance obviously.

\noindent \textbf{Training tasks.}
We also perform a drop-one-out ablation on the multi-modal multi-task training data, in \cref{tab:ablation_tasks}.
\textbf{Detection} data provides samples for localization, especially multi-instance situation.
It is instinctively important for learning to localize, and indeed matters for performance.
\textbf{I2T} (image-to-text, like image captioning and visual question answering) often helps the generalization and zero-shot performance of multi-modal methods. We find that it does affect the zero-shot performance on \ddd{} greatly.
\textbf{MLM} is theoretically important for language understanding and generalization. However, we find it actually is not.
Removing the MLM task has no significant effect on the performance.
We surmise that the generalization ability of OFA-DOD on \ddd{} mainly comes from I2T.

\section{Conclusion and Limitation}
\label{sec:conclusion}

% In this paper, we present a more flexible and realistic setting for object understanding, called Described Object Detection, which encompasses both object detection and referring expression comprehension tasks. For this, we propose a benchmark more challenging than existing benchmarks, with unrestricted language reference, complete annotation, and absence expressions. We evaluate several state-of-the-art methods from previous tasks on this dataset, and propose a new baseline outperforming them for future research.
% This work contributes to the development of object understanding and provides a new benchmark for evaluating models in this area.

In this paper, we bring the Described Object Detection (DOD) task to the foreground. 
For this task, we introduce a dataset called \ddd{}, which annotates described objects without omission and features flexible language expressions, whether long or short, complex or simple. 
Our evaluation of SOTA methods from REC or OVD on \ddd{} reveals challenges faced by REC, OVD, and bi-functional approaches.
Based on these observations, we propose a baseline that largely improves REC methods for DOD task.
We believe that the dataset and findings will contribute to advancing the understanding and development of DOD methods, facilitating future research in this area.

\noindent \textbf{Limitation and broader impact.}
% This work establishes the research foundation for the DOD task. Compared to traditional detection algorithms, DOD models have lower customization thresholds, enabling users to specify the detection target using language. This may lead to potential abuse, such as detecting people's privacy. Moreover, a significant amount of GPU computing was utilized in model development and algorithm evaluation, resulting in carbon emissions.
This work does have some limitations. Due to the significant annotation cost brought by our complete annotation process, we are unable to propose a huge dataset with millions or billions of images. Besides, the evaluation and findings in this work may be dependent on the choice of descriptions and the image sources. This work only serves as a starting point for DOD and we hope there will be other DOD datasets with larger scales.
In the broader community, compared to traditional detection algorithms, DOD models have a lower customization threshold, enabling users to specify the detection target using language. This may lead to potential abuse.

\noindent \textbf{Future work.}
During peer-review process, some new works with potential for DOD emerges, including Shikra~\cite{chen2023shikra}, Kosmos-2~\cite{peng2023kosmos} and Qwen-VL~\cite{bai2023qwenvl}. We will continue to investigate such methods for DOD and update them in \href{https://github.com/Charles-Xie/awesome-described-object-detection}{this list}.

\noindent \textbf{Acknowledgments.}
This work was supported in part by the National Natural Science Foundation of China under Grant 62076183, 61936014 and 61976159, in part by the Natural Science Foundation of Shanghai under Grant 20ZR1473500, in part by the Shanghai Science and Technology Innovation Action Project under Grant 20511100700 and 22511105300, in part by the Shanghai Municipal Science and Technology Major Project under Grant 2021SHZDZX0100, and in part by the Fundamental Research Funds for the Central Universities. The authors would also like to thank the anonymous reviewers for their careful work and valuable suggestions.

{
\small
\bibliographystyle{abbrv}
\bibliography{nips_bib}
}

%%%%%%%%%%%%%%%%%%%%%%%%%%%%%%%%%%%%%%%%%%%%%%%%%%%%%%%%%%%%

\clearpage

\appendix

\centerline{\Large{\textbf{Described Object Detection: Liberating Object Detection with Flexible Expressions}}} 
\vspace{10pt}
\centerline{\large{\textit{{\rule[0.25\baselineskip]{0.12\textwidth}{1pt}\ \ Supplemental File\ \ \rule[0.25\baselineskip]{0.12\textwidth}{1pt}}}}} 

\vspace{5pt}
\begin{abstract}
{\noindent \bf Content.} In this supplemental file, we provide more details of this work to supply the main paper.
    \begin{itemize}
        \item[$\blacktriangleright$] \textbf{Dataset details and more examples} for the proposed \ddd{} dataset are presented in \cref{sec:dataset_details}.
        \item[$\blacktriangleright$] \textbf{Evaluation of previous methods} are presented in \cref{sec:eval_baselines}, which describes the existing works we evaluated and the specific details regarding how we adapt them to the DOD task.
        \item[$\blacktriangleright$] \textbf{Details of the proposed baseline} are presented in \cref{sec:method_details}.
        \item[$\blacktriangleright$] \textbf{More experimental results} are shown in \cref{sec:additional_exp}, including both quantitative and qualitative results.
    \end{itemize}    
\end{abstract}

\section{Dataset Details}
\label{sec:dataset_details}

\subsection{More examples}

In the Section 3.1 of the main paper, we introduce the characteristics of the proposed \ddd{} dataset, and elaborate the 3 major ones in Fig. 2 with some examples. Here we provide more examples to supplement this part.

\begin{figure}[t]
\begin{center}
   \includegraphics[width=\linewidth]{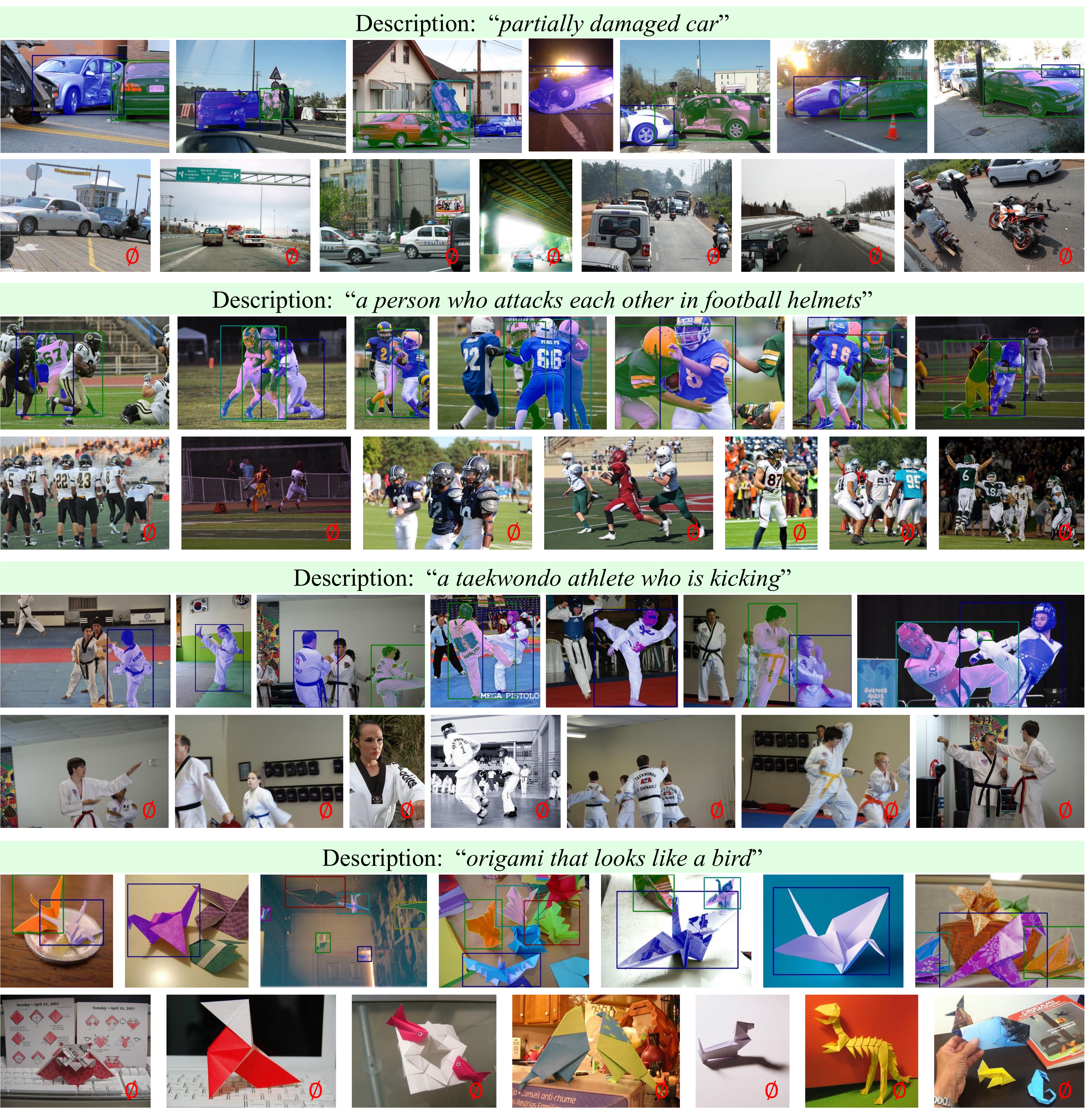}
    % \vspace{-25pt}
\end{center}
   \caption{
   Examples demonstrate that the proposed \ddd{} is fully annotated with positive and negative examples across the entire dataset. The visualizations include four descriptions along with selected positive and negative image samples from the dataset. Each description is accompanied by two rows of image samples: the first row contains positive images, and the second row contains negative images. For positive images, the specific description's bounding boxes and instance masks are visualized. In contrast, for negative images, an empty set symbol \textcolor{red}{$\emptyset$} is displayed in red at the right corner.
   The visualizations are best observed in color and with zoomed-in view.
   }
\label{fig:examples_completeness}
\end{figure}

\noindent \textbf{Complete annotation.}
The first characteristic of \ddd{} is the dataset-level complete and thorough annotations, setting it apart from REC datasets~\cite{yu2016modeling,mao2016generation}. In \ddd{}, every image is annotated for possible positive and negative instances, as demonstrated in \cref{fig:examples_completeness}. This figure includes several images with positive instance labels (first row) and several images with negative instance labels (second row) for each of the four descriptions. Such comprehensive annotation makes the proposed dataset well-suited for detection tasks.

In comparison, REC datasets like RefCOCO~\cite{yu2016modeling,mao2016generation} only annotate several positive instances in a few images for each description, leaving all the other images without annotations for that particular description; thus, their annotation completeness is limited to the image-level. On the other hand, GRD~\cite{wu2023gres} annotates a description for a group of images while dividing the entire set into multiple groups, resulting in an annotation completeness at the group-level.

\begin{figure}[t]
\begin{center}
   \includegraphics[width=\linewidth]{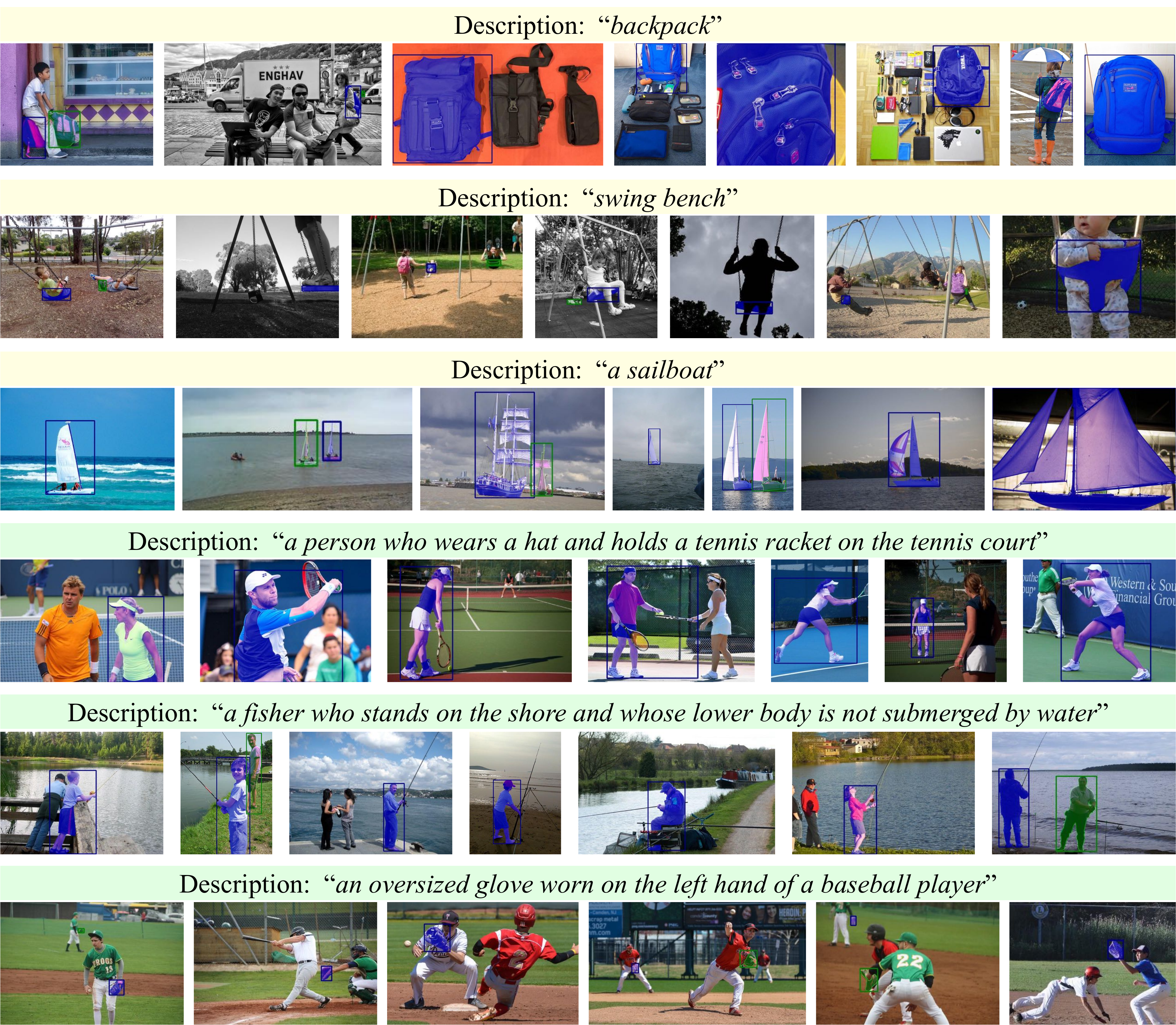}
    % \vspace{-25pt}
\end{center}
   \caption{Examples showing the descriptions in \ddd{} are free-form and unrestricted. The descriptions can be short and simple (like the top 3 descriptions, in yellow background) or long and complex (like the bottom 3, in green background).
   Boxes and instance masks belonging to the specific description are visualized in each image.
   The visualizations are best observed in color and with zoomed-in view.
   }
\label{fig:examples_freeness}
\end{figure}

\noindent \textbf{Unrestricted description.}
The categories in \ddd{} encompass more than just simple object names, such as \texttt{cat}, \texttt{dog} and \texttt{bird} found in typical object detection datasets~\cite{lin2014microsoft,gupta2019lvis,shao2019objects365}. As illustrated in \cref{fig:examples_freeness}, the descriptions are expressed in unrestricted natural language. The longer and more complex descriptions resemble references found in REC datasets~\cite{yu2016modeling,mao2016generation,kazemzadeh2014referitgame}. For instance, a description like \texttt{a fisher who stands on the shore and whose lower body is not submerged by water} comprises 16 words and encompasses multiple attributes like \texttt{fisher}, \texttt{stands on the shore} and \texttt{lower body is not submerged by water}. These attributes are semantically abstract and visually diverse. On the other hand, the shorter and simpler descriptions can be similar to the category names in OD datasets, such as \texttt{backpack}, \texttt{swing bench} and \texttt{a sailboat}.
This illustrates that the descriptions of objects in \ddd{} are free-form and unrestricted, covering a wide range of description types present in both REC and OD datasets.

\begin{figure}[t]
\begin{center}
   \includegraphics[width=\linewidth]{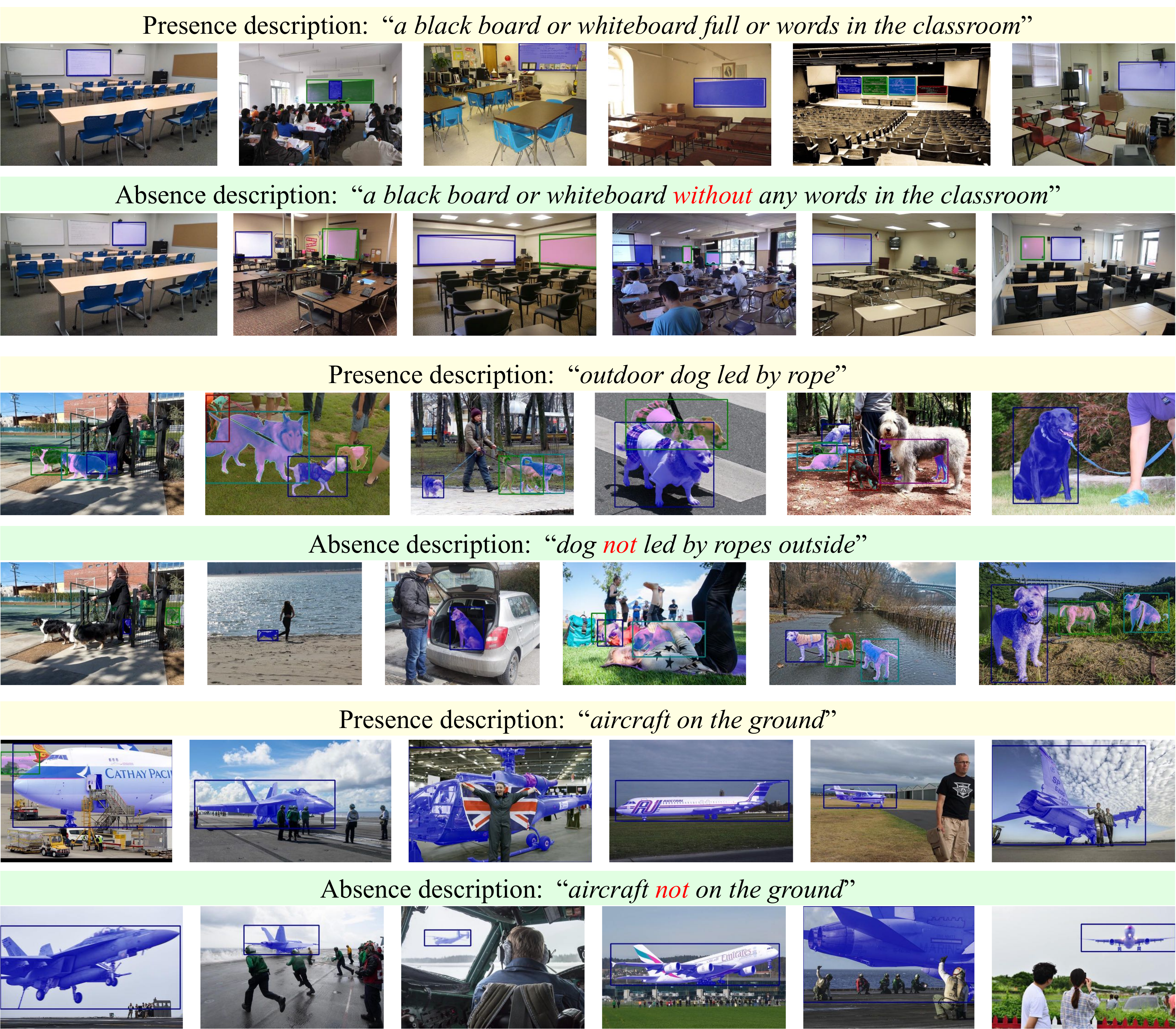}
    % \vspace{-25pt}
\end{center}
   \caption{
   Examples showing the presence and absence descriptions in \ddd{}.
   Six descriptions, containing 3 pairs of contrary presence descriptions (in yellow background) and absence descriptions (in green background), are illustrated alongside their corresponding positive examples. The key words depicting absence expressions are in red. Boxes and instance masks belonging to the specific description are visualized in each image.
   The visualizations are best observed in color and with zoomed-in view.
   }
\label{fig:examples_absence}
\end{figure}

\noindent \textbf{Absence description.}
To the best of our knowledge, the proposed dataset is the first annotated dataset specifically designed to address absence descriptions. Examples with annotations for both presence and absence descriptions from our dataset (\ddd{}) are illustrated in \cref{fig:examples_absence}. For visualization purposes, we have selected some absence descriptions that have contradictory presence descriptions. The absence descriptions and the corresponding presence descriptions differ primarily in the existence of key attributes. For instance, the first presence description emphasizes black/white boards \textit{with} words written, while the first absence description focuses on those \textit{without} words.

It is important to note that in certain cases, some images contain both absence and presence descriptions. For example, in the first example image of the second presence-absence pair, both dogs led by ropes and not led by ropes coexist. Such instances pose significant challenges, as they require the DOD model to comprehend the absence of concepts in a language description and to discern the subtle differences among instances within an image.

\noindent \textbf{Other characteristics for instance annotations.}
Examples in \cref{fig:examples_completeness,fig:examples_freeness,fig:examples_absence} all illustrate some additional characteristics of \ddd{}:

(1) Instance-level annotation, where each instance is individually labeled.
(2) One description can refer to multiple instances in an image.
(3) Each instance is annotated with both bounding boxes and masks. As a result, the proposed dataset is not limited to the Described Object Detection setting focused on in this work but can also support a similar task, producing instance segmentation masks rather than object detection bounding boxes.

\subsection{More statistics}

The proposed dataset contains a total of 10,578 images, 18,514 boxes (including instance masks), and 422 well-designed descriptions. These descriptions comprise 316 presence descriptions and 106 absence descriptions.

Regarding the inter-scenario setting, considering all 422 descriptions, there are 24,282 positive object-text pairs and 7,788,626 negative pairs. When considering only positive descriptions, there are 16,480 positive pairs and 5,833,944 negative pairs.

For the intra-scenario setting (where candidate descriptions for an image only come from the same scenario), there are 20,279 positive pairs and 53,383 negative pairs. For the subset with only positive descriptions, there are 13,917 positive pairs and 41,231 negative pairs.

The average expression length in the dataset is 6.3 words.

\begin{figure}[t]
\begin{center}
    \begin{subfigure}[b]{0.40\textwidth}
     \centering
     \includegraphics[width=\textwidth]{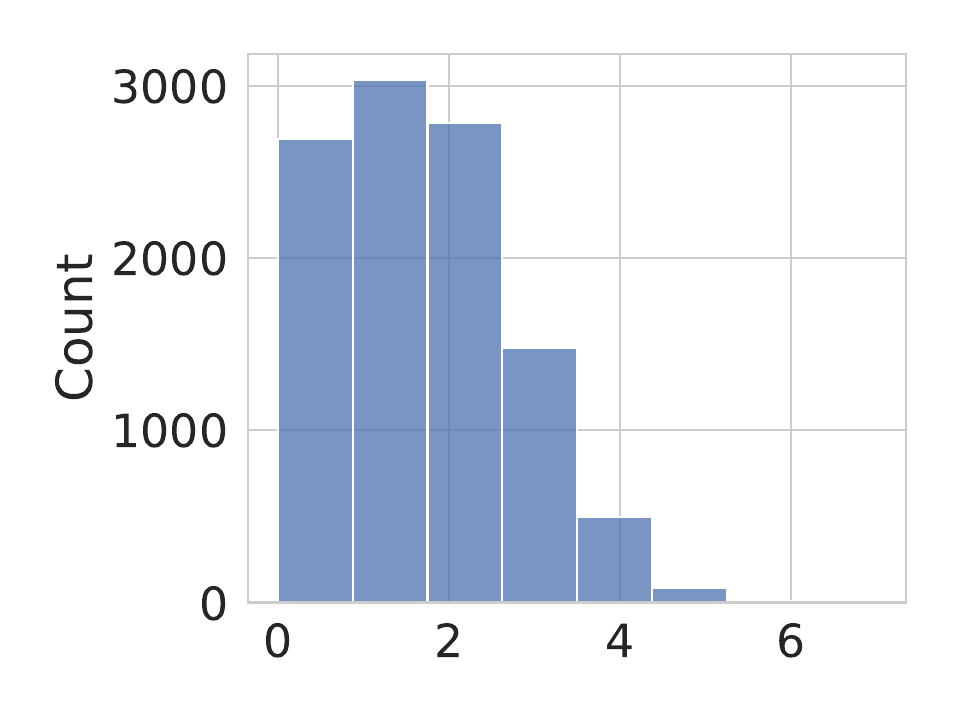}
     \caption{
     Distribution of number of descriptions on one image.
     }
     \label{fig:nref_per_img}
    \end{subfigure}
    \hfill
    \begin{subfigure}[b]{0.40\textwidth}
     \centering
     \includegraphics[width=\textwidth]{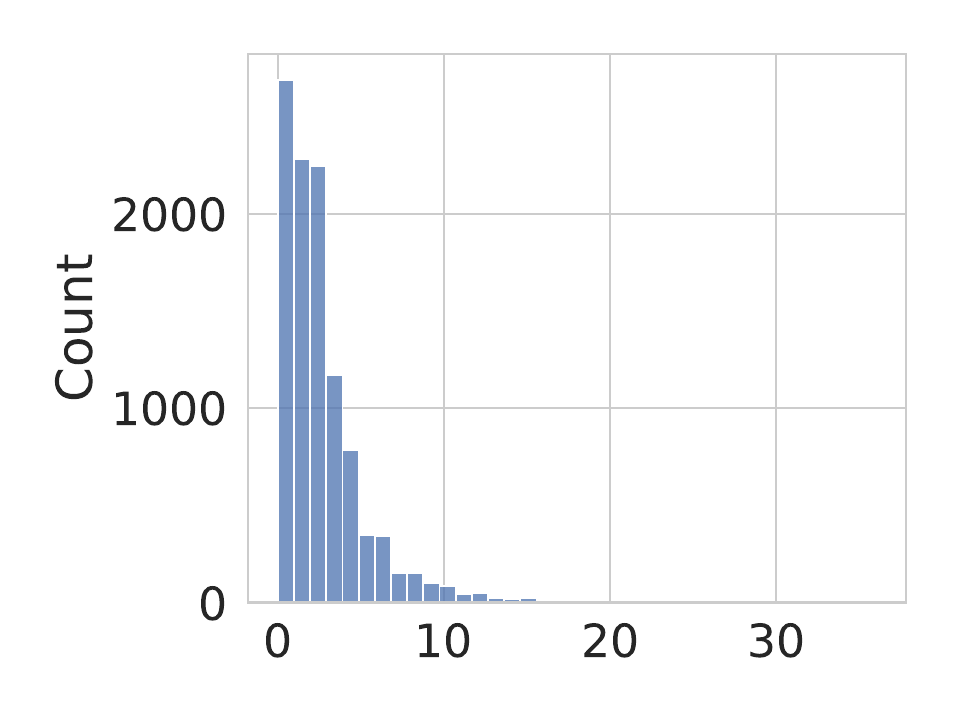}
     \caption{
     Distribution of number of instances on one image.
     }
     \label{fig:nbox_per_img}
    \end{subfigure}
\end{center}
   \caption{Distribution of (a) number of positive descriptions on an image in the dataset, and (b) number of positive instances on an image in the dataset. (a) shows that the majority of images contains multiple positive descriptions in the proposed dataset, while (b) shows that many images contains multiple boxes.}
\label{fig:lengths_nums_supp}
\end{figure}

In \cref{fig:lengths_nums_supp}, two additional histograms demonstrate the distribution of the number of positive descriptions and the number of positive instances within a single image in the dataset. This visualization highlights the complexity of the proposed dataset, with frequent occurrences of multiple references and many instances within one image.

\noindent \textbf{Absence descriptions.}
To the best of our knowledge, the proposed \ddd{} benchmark is the first to investigate the capability of models to comprehend the absence of certain features and attributes and distinguish between absence and presence. This unique focus on absence-related comprehension sets it apart from previous benchmarks with description annotation (e.g., datasets like RefCOCO~\cite{yu2016modeling,mao2016generation} for REC and RES tasks). Notably, RefCOCO contains an extremely small and neglectable number of instances with absence descriptions. In contrast, the \ddd{} dataset comprises 106 absence expressions out of a total of 422 descriptions, approximately 25\%, and 7,802 positive annotated instances. This significant inclusion of absence-related expressions contributes to a vital and distinguishing characteristic of our proposed benchmark.

\noindent \textbf{Category overlapping with previous datasets.}
The proposed dataset can be regarded as an OVD benchmark (but with longer references rather than category names), if we take classes and references in previous OVD/REC datasets as \textit{base} classes, and the classes in \ddd{} as \textit{novel}.
Categories in \ddd{} has very little overlap with previous datasets. Here we try to quantify the minimal overlap between \textit{base} (OVD datasets like COCO/LVIS and REC datasets like RefCOCO/+/g) and \textit{novel} ($D^3$).
For comparison with OVD datasets, we used ChatGPT to generate synonyms from category names in those datasets and then match them against references in $D^3$. The overlapping percentage is 0.4\% for COCO and 0.9\% for LVIS.
For COCO, which have less categories, we also perform manual check and calculation, resulting in 0.7\% overlap with $D^3$.
For REC datasets, we apply a threshold on the sentence similarity calculated via HuggingFace's \texttt{bert-base-cased-finetuned-mrpc} model. The calculated overlaps of $D^3$ with RefCOCO/+/g is 0.0\%, 0.2\% and 0.7\%, separately.
Thus, novel classes ($D^3$) overlap <1\% with base classes (from OVD \& REC datasets).

\subsection{Annotation process}

\begin{figure}[t]
\begin{center}
   \includegraphics[width=\linewidth]{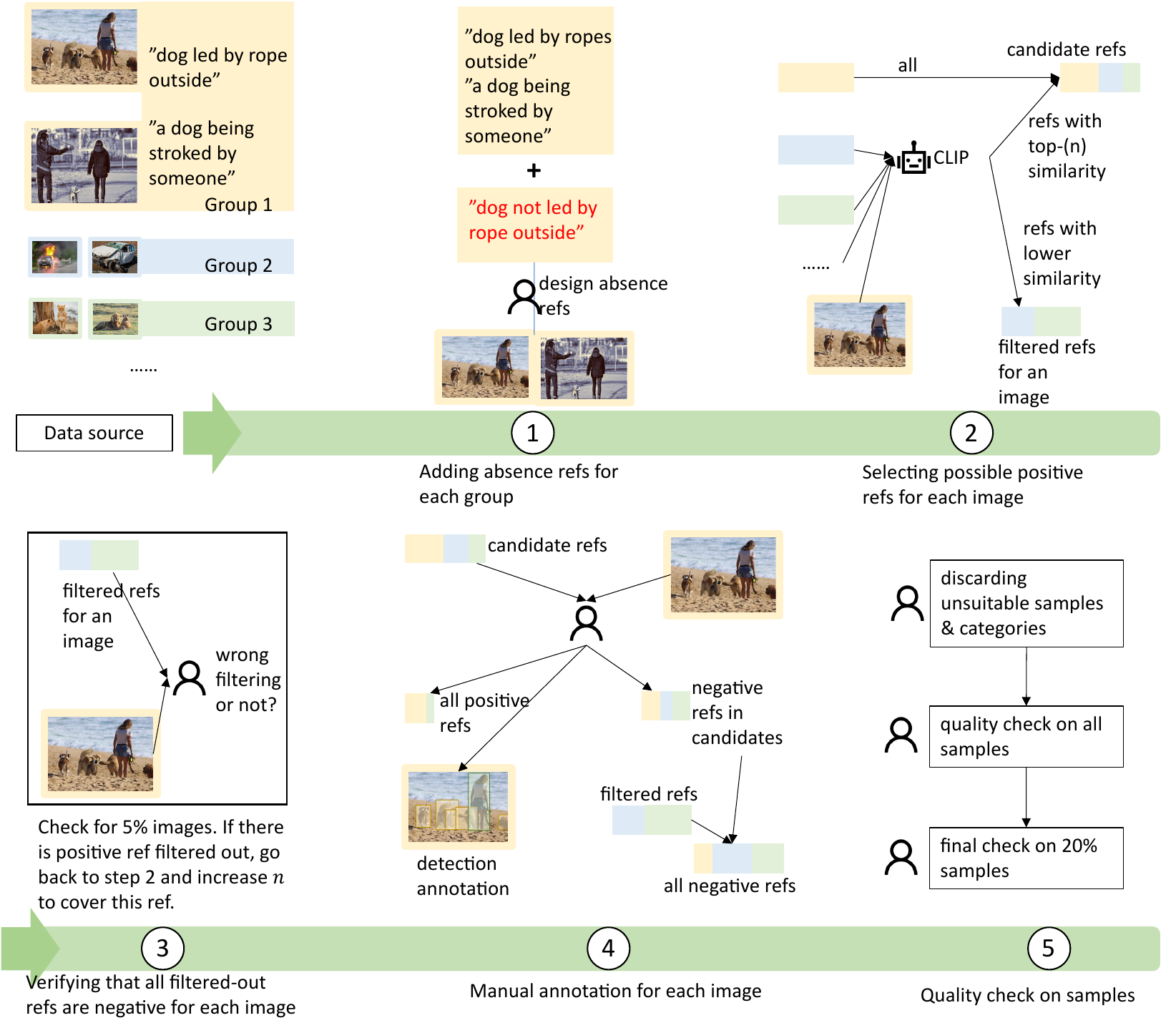}
\end{center}
   \caption{Annotation process of the proposed $D^3$ benchmark.
   }
\label{fig:annotation_process}
\end{figure}

The data source of \ddd{] is 106 groups from GRD~\cite{wu2023gres}, with about 100 images crawled from \href{https://www.flickr.com/}{Flickr} and 3\textasciitilde 4 designed refs for each group. Each group belongs to a different scenario and the overlapping between refs from different groups are small (i.e, a ref for one group are not frequent (but possible) to appear in the images from another group). Now we have 10000+ images and 300+ refs.

A diagram illustrating the annotation process of \ddd{} is presented in \cref{fig:annotation_process}. Here we describe the details of the annotation steps as below:

\begin{enumerate}
    \item \textsc{Manual} Adding absence refs: design 1\textasciitilde 2 absence refs based on the images for each group and add them to the corresponding groups. Now we have 400+ refs.
    \item \textsc{Automatic} Selecting possible positive refs: for each image, select \textit{all the refs} (4\textasciitilde 6) from the group it belongs to, and also the other 105 groups (top-$n$ refs out of 400+ refs, by CLIP similarity between the image and each description). Now for each image, we have $n+4$\textasciitilde $n+6$ candidate refs and all the other refs are filtered out. $n$ is set as 40 initially.
    \item \textsc{Manual} Verification: randomly choose 5 groups of images, and check if there are any positive refs that should not be filtered out. If so, increase $n$ to cover that ref and go back to step 2.
    \item \textsc{Manual} Human annotation: annotation by trained annotators on all images. The annotation of boxes (and instance masks) are instance-level, dataset-wise complete, and includes absence refs.
    \item \textsc{Manual} Quality check: this includes 3 small steps:
    \begin{enumerate}
        \item Discarding some images (unsuitable for annotation, e.g., ambiguity) or categories from the dataset. About 8\% samples are discarded.
        \item Quality check on 100\% samples. For each group, if image with error is more than 2\%, it is returned for re-annotation. Otherwise the errors are fixed and this group passes this step.
        \item Final check on 5\% samples. For each group, if there are image with error, it is returned, otherwise it is accepted.
    \end{enumerate}
\end{enumerate}

\section{Evaluating Existing Baselines}
\label{sec:eval_baselines}

In Section 4.1 of the paper we evaluate several representative and SOTA methods for OVD~\cite{minderer2022simpleOWLViT,wu2023cora}, REC~\cite{wang2022ofa} and bi-functional methods~\cite{liu2023groundingdino,yan2023universal} on the proposed \ddd{} for the DOD task.
Here we introduce these methods and describe how we adapt them to DOD and evaluate them on \ddd{}. Notably, the images in \ddd{} do not overlap with the training data of these existing baselines and our proposed baseline, so all the comparisons are actually conducted under zero-shot setting, and is relatively fair.

\noindent \textbf{OFA.}
OFA is the SOTA REC method. It is proposed as a general-purpose vision-language model, with ability to performing various tasks like image captioning (IC), VQA, referring expression comprehension (REC), etc.
It adopts data from various tasks for pretraining, including MLM, IC, VQA, REC, and OD.
Notably, through pretrained on object detection datasets~\cite{lin2014microsoft,gupta2019lvis}, it is not evaluated on these tasks at all. We find that a pretrained OFA model merely achieves 9.6 mAP on COCO~\cite{lin2014microsoft} benchmark, which is too far from modern object detectors. This is also the reason we do not include it as bi-functional models.

OFA can be evaluated on a downstream task either after pretraining or after fine-tuning on the specfic dataset.
On REC datasets, it is already strong with only pretraining and achieves SOTA performance after fine-tuning on REC only.
As the images in \ddd{} do not overlap with those in REC datasets, we use the pretrained model of OFA rather than the one fine-tuned on REC data, for better generalization ability.
The official checkpoints are used as the model to evaluate on \ddd{}. Model checkpoints of multiple sizes are available and we use the largest two, namely OFA-base and OFA-large.

For REC task, OFA takes in a pair of one image and one sentence, and predicts a sequence of 4 coordinates, which forms a bounding box.
For DOD, we apply a similar inference strategy. For a image and the candidate descriptions (for intra-scenario setting, only a few descriptions in that scenario; for inter-scenario setting, all the descriptions in the dataset), each description and the image form a input image-text pair and predicts a detected instance (bounding box) that will be saved as the result.
As OFA predicts token sequences of box coordinates and no classification scores, we use the average of the classification score on the 4 coordinate tokens as the confidence score for each detected instance. No further processing is applied.

\noindent \textbf{OWL-ViT.}
OWL-ViT~\cite{minderer2022simpleOWLViT} and CORA~\cite{wu2023cora} are the SOTA OVD methods.
OWL-ViT also adopts a pretraining and fine-tuning strategy for training. It is pretrained with image-text contrastive learning, similar to CLIP~\cite{radford2021learning} and then transferred to OVD with simple modification and fine-tuning on standard detection datasets. For evaluation on \ddd{}, we use the model fine-tuned on detection datasets without other training.
Model checkpoints with ViT-base~\cite{dosovitskiy2021image} and ViT-large backbones are available.

For OVD, OWL-ViT takes in some text sequences and one image, and predicts a lot of instances consisting of bounding boxes, class labels as well as classification scores. The text sequences are category names like \texttt{giraffe}, \texttt{car}, etc. The detected instances with a score less than threshold 0.1 are filtered.
For the proposed DOD, we apply a similar inference strategy. The input text is the candidate descriptions, and the output instances are filtered by the same threshold 0.1. No other modifications or post-process are applied.

\noindent \textbf{CORA.}
CORA~\cite{wu2023cora} is a DETR~\cite{kamath2021mdetr} style method that adapts CLIP~\cite{radford2021learning} to OVD. It takes CLIP as the pretrained model and fine-tune the modified framework on detection datasets~\cite{lin2014microsoft,gupta2019lvis}.

The inference of CORA on OVD is performed as a matching between image region features and category name embeddings encoded by CLIP text encoder. For inference on DOD, we adopt the same strategy. We only replace the input images with the images from \ddd{} and the category names with the candidate descriptions. Other details follow the settings in CORA for OVD.

\noindent \textbf{Grounding-DINO.}
The bi-functional Grounding-DINO~\cite{liu2023groundingdino} extends a close-set object detector DINO~\cite{zhang2023dino} to open-set object detection. It is pretrained on vast object detection~\cite{lin2014microsoft,gupta2019lvis,Krasin2017,shao2019objects365} and image captioning data~\cite{sharma2018conceptual,thomee2016yfcc100m,ordonez2011im2text}. However, this model is not competitive on REC, and a further fine-tuning on REC data~\cite{yu2016modeling,mao2016generation}is required to achieve a strong performance.
Official model checkpoints with Swin-tiny~\cite{liu2021swin} and Swin-base backbones are available.

It produces a lot of detected instances for one image-text input, and filters some instances with a threshold hyper-parameter. For the inference on REC, given an image-reference pair, it merely keeps the one and only instance with the largest score.
We follow its inference process on REC task for the proposed DOD. We will dig more into the specific inference strategy and hyper-parameters in the additional experiments in \cref{sec:additional_exp}. 

\noindent \textbf{UNINEXT.}
UNINEXT~\cite{yan2023universal} stands as another bi-functional method, reformulating a diverse array of tasks, such as object detection, REC, video-based tracking, image and video segmentation tasks, into a unified multi-task framework that excels in instance prediction and retrieval. This innovative approach involves three stages of pre-training without any single-task fine-tuning. In the first stage, training is performed with Object365~\cite{shao2019objects365}, followed by the second stage with REC data and COCO, and finally, the third stage with extensive data from video tasks.

For evaluation on \ddd{}, we utilize the UNINEXT models trained in the second stage, which only utilizes image data and is relatively fair for comparison. Model checkpoints featuring ConvNeXt-large and ViT-huge backbones are available, and these are the ones we employ for evaluation.

For each task it is pretrained on, UNINEXT designs an individual inference strategy.
For the DOD task, we adopt an inference strategy similar to REC. To delve deeper into the specific inference strategy and hyper-parameters, we also conduct additional experiments in \cref{sec:additional_exp}.

\section{The Proposed Baseline}
\label{sec:method_details}

As stated in Section 4.2 of our paper, we choose OFA as the foundation for the proposed baseline. Here we provides two figures to show the differences between OFA~\cite{wang2022ofa} in~\cref{fig:ofa_structure} and the proposed OFA-DOD in~\cref{fig:ofa_dod_structure}. 

As shown in the two figures, the first modification, granularity decomposition, corresponds to replacing a shared decoder with two parallel decoders, one for global tasks and one for local tasks; the second modification, reconstructed data, refers to the reconstructed OVD \& REC data for the local decoder, after which the input can be one or multiple references (or object category names) and they can corresponds to zero, one or multiple targets; the third modification, task decomposition, is depicted by adding a binary classification in the global decoder, which determines if a bounding box and a description is matched.

\begin{figure}[t]
\begin{center}
   \includegraphics[height=3.5cm]{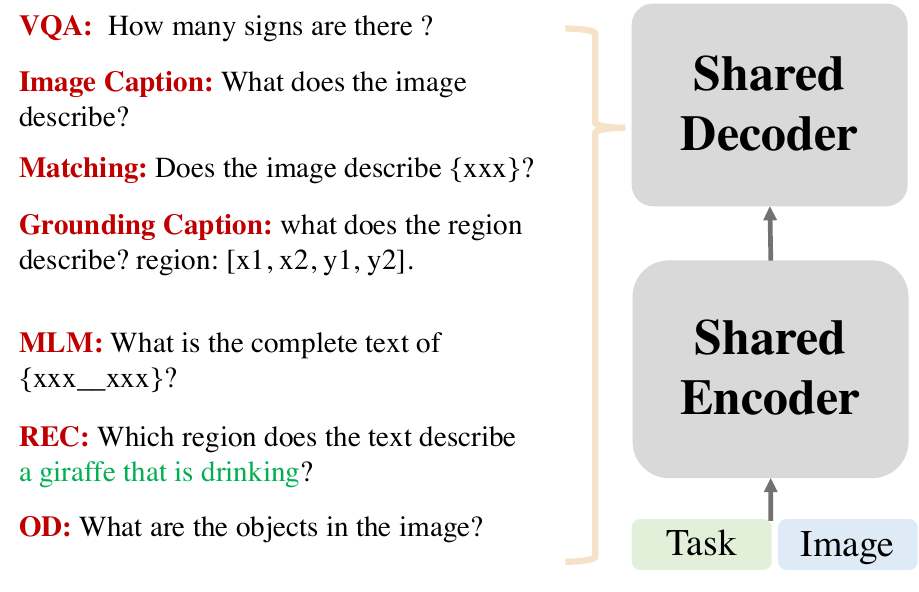}
\end{center}
   \caption{Model structure of OFA~\cite{wang2022ofa}.
   }
\label{fig:ofa_structure}
\end{figure}

\begin{figure}[t]
\begin{center}
   \includegraphics[height=3.5cm]{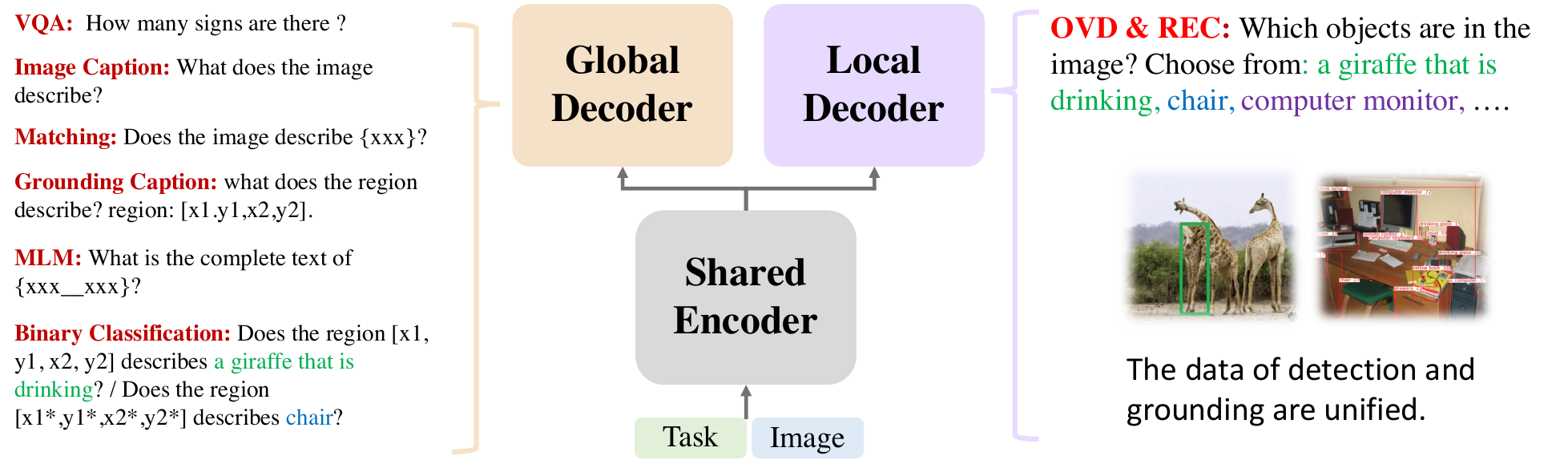}
\end{center}
   \caption{Model structure of the proposed OFA-DOD.
   }
\label{fig:ofa_dod_structure}
\end{figure}

More details regarding these 3 modifications are stated below:

\subsection{Granularity decomposition}

The aim of this adjustment is to enhance the suitability of the baseline for localization tasks such as OVD, REC, and DOD.
The original OFA~\cite{wang2022ofa} consists of a multi-modal encoder and a decoder.
For each task, whether it involves image-only, text-only, or image-text inputs, an image (which can be omitted) and a text prompt are fed into the multi-modal encoder to predict the output as a text sequence. 
All task processes are forced to co-exist in one encoder and one decoder.

To achieve this decomposition, we divide the pretraining tasks of OFA into two different granularities: global tasks for language modeling-related tasks like IC, VQA, MLM, etc., and local tasks for region localization-related tasks such as object detection and REC.
We add an extra decoder alongside the original one, which also takes input from the encoder. The two decoders handle global and local tasks independently, thereby avoiding mutual interference.

This improvement effectively resolves conflicts between different tasks and enhances the capability of the model for localization tasks.

\subsection{Reconstructed data}

This improvement is to benefit detection with multiple target instances.
For OFA, REC is performed with one image and one text prompt (question prefix concatenated with one description) as input, and a bounding box sequence with 4 coordinate tokens as output. The input sequence has the form:

\begin{center}
    \textcolor{blue}{Which region does the text [REF1] describe?} \textcolor{green}{[IMG1]},
\end{center}

where \textcolor{blue}{[REF1]} is a description annotated for the image, and \textcolor{green}{[IMG1]} is the image token sequence.

Originally, each input example in REC is a image-text-box pair, where one reference is annotated with one bounding box for one image.
We reconstruct the data of REC by 2 steps:
First, we grouping the descriptions belonging to one image, and each reconstructed input example is a combination of one image, $N$ positive descriptions, and $N$ boxes, where $N$ is a integer equal to or larger than 1.
Second, for each image, we sample some descriptions from other images as the negative description.
With the prepared data, we change the input as:

\begin{center}
    \textcolor{blue}{Which of these options are in the image? Choose from options: [REF1] [REF2] [REF3] ...}
    \textcolor{green}{[IMG1]},
\end{center}

where \textcolor{blue}{[REF1] [REF2] [REF3]} are positive or negative randomly sampled.
The output is to predict a series of multiple boxes, each followed by its corresponding descriptions in the input.
This results in a unified data format for OD and REC. For OD, the negative descriptions are negative class names.
The reformulated data are noisy, as they are not initially prepared for DOD, and a sampled negative description is not necessarily negative due to the image-level annotation completeness of REC. Still, we find such reconstructed data helpful.

\subsection{Task decomposition}

This step aims to enhance the baseline's capability to discern false positives. In addition to training on REC (to locate a region based on a reference), we leverage the multi-task nature of OFA by introducing an additional VQA task. This task involves determining whether a predicted region and a description match with each other and can be viewed as a binary classification problem.
The input for this VQA task is:

\begin{center}
    \textcolor{blue}{Does the region [BOX1] describes [REF1]?}
    \textcolor{green}{IMG1},
\end{center}

where \textcolor{blue}{[BOX1]} is the bounding box coordinate tokens corresponds to the description. For training, the box and the reference are either from a GT text-box pair, or the GT box is shifted (as negative sample), or the box and the reference are from different text-box pairs (as negative sample, too).
The output of this task is a text sequence \textcolor{blue}{yes} for positive samples and \textcolor{blue}{no} for negative samples.
This step is responsible for rejecting possible false positives.

\section{More experimental results}
\label{sec:additional_exp}

\begin{table}[t]
    \centering
    \caption{Comparison of different methods on the proposed dataset for different mAP metrics: intra-secnario mAPs, inter-scenario mAPs, and average recalls. ``Bi'' denotes bi-functional methods.}
    \resizebox{\textwidth}{!}{
    \begin{tabular}{l l | c c c | c c c | c c c }
        \hline
        \multirow{2}{*}{Task} & \multirow{2}{*}{Method} & \multicolumn{3}{c}{Intra-scenario} & \multicolumn{3}{| c}{Inter-scenario} & \multicolumn{3}{c}{Average Recall} \\
        & & \textsl{FULL} & \textsl{PRES} & \textsl{ABS} & \textsl{FULL} & \textsl{PRES} & \textsl{ABS} & \textsl{FULL} & \textsl{PRES} & \textsl{ABS} \\
        \hline
        \multirow{2}{*}{REC} & OFA$_\text{base}$ & 3.4 & 3.0 & 4.3 & 0.1 & 0.1 & 0.1 & 13.7 & 13.5 & 14.3 \\
        & OFA$_\text{large}$ & 4.2 & 4.1 & 4.6 & 0.1 & 0.1 & 0.1 & 17.1 & 16.7 & 18.4 \\
        \hline
        \multirow{3}{*}{OVD} & CORA$_\text{R50}$ & 6.2 & 6.7 & 5.0 & 2.0 & 2.2 & 1.3 & 10.0 & 10.5 & 8.7 \\
        & OWL-ViT$_\text{base}$ & 8.6 & 8.5 & 8.8 & 3.2 & 3.7 & \textbf{4.7} & 13.5 & 13.7 & 13.1 \\
        & OWL-ViT$_\text{large}$ & 9.6 & 10.7 & 6.4 & 2.5 & 2.9 & 2.1 & 17.5 & 19.4 & 11.8 \\
        \hline
        \multirow{4}{*}{Bi} & UNINEXT$_\text{large}$ & 17.9 & 18.6 & 15.9 & 2.9 & 3.1 & 2.5 & 40.7 & 42.6 & 34.7 \\
        & UNINEXT$_\text{huge}$ & 20.0 & 20.6 & 18.1 & 3.3 & 3.9 & 1.6 & 45.3 & 46.7 & 41.4 \\
        & G-DINO$_\text{tiny}$ & 19.2 & 18.5 & 21.2 & 2.3 & 2.5 & 2.1 & 47.8 & 48.1 & 46.6 \\
        & G-DINO$_\text{base}$ & 20.7 & 20.1 & \textbf{22.5} & 2.7 & 2.4 & 3.5 & 51.1 & 51.8 & 48.9 \\
        \hline
        DOD & OFA-DOD$_\text{base}$ & \textbf{21.6} & \textbf{23.7} & 15.4 & \textbf{5.7} & \textbf{6.9} & 2.3 & 47.4 & 49.5 & 41.2 \\
        \hline
    \end{tabular}
    }
    \label{tab:comparison_supp}
\end{table}

\begin{table}[t]
    \centering
    \caption{Performance of bi-functional methods~\cite{liu2023groundingdino,yan2023universal}, compared with the proposed baseline, under different score filtering thresholds. The mAP under \textsl{FULL} setting and the False Positive Per Category (FPPC) on images with no instance for one category are reported as metrics. For methods filtered with different score thresholds, we highlight the rows when they achieve a FPPC similar to our OFA-DOD.}
    \begin{tabular}{l | c | c c}
        \hline
        \multirow{2}{*}{Method} & \multirow{2}{*}{Threshold} & No-instance & \textsl{FULL} \\
        & & FPPC (\%) $\downarrow$ & mAP (\%) $\uparrow$ \\
        \hline
        \multirow{8}{*}{UNINEXT~\cite{yan2023universal}} & - & 100.0 & 20.0 \\
        % & 0.2 & 100.0 & 20.0 \\
        % & 0.3 & 99.9 & 20.0 \\
        & 0.4 & 99.3 & 20.0 \\
        & 0.5 & 96.5 & 19.9 \\
        & 0.6 & 84.0 & 19.7 \\
        & 0.7 & 57.8 & 18.1 \\
        & \textbf{0.8} & \textbf{36.0} & \textbf{15.7} \\
        & 0.9 & 11.5 & 8.7 \\
        \hline
        \multirow{8}{*}{Grounding-DINO~\cite{liu2023groundingdino}} & - & 100.0 & 20.7 \\
        % & 0.2 & 99.8 & 20.7 \\
        % & 0.3 & 96.0 & 20.7 \\
        & 0.4 & 80.8 & 20.2 \\
        & 0.5 & 60.6 & 18.4 \\
        & 0.6 & 45.2 & 16.2 \\
        & \textbf{0.7} & \textbf{34.6} & \textbf{13.6} \\
        & 0.8 & 23.3 & 9.5 \\
        & 0.9 & 8.5 & 3.8 \\
        \hline
        OFA-DOD & - & \textbf{35.6} & \textbf{21.6} \\
        \hline
    \end{tabular}
    \label{tab:binfunctional_thres}
\end{table}

\begin{figure}[t]
\begin{center}
   \includegraphics[width=\linewidth]{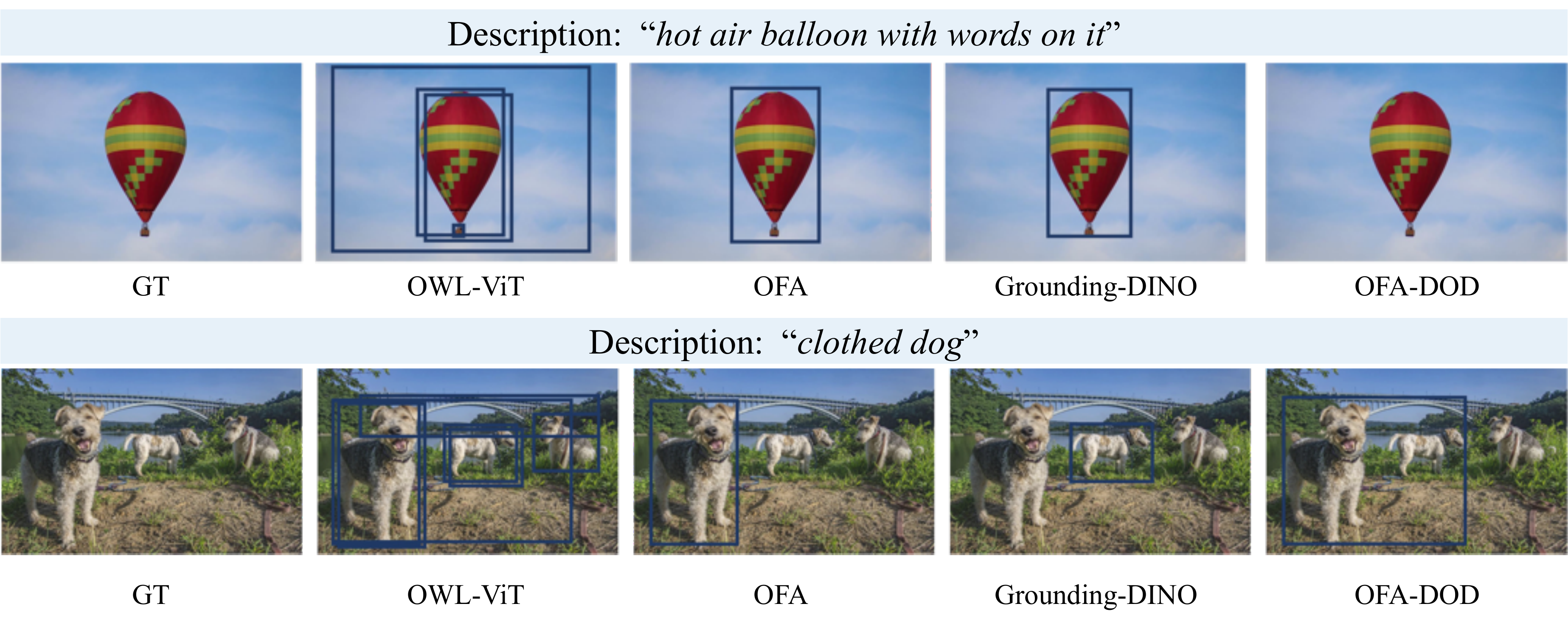}
    % \vspace{-25pt}
\end{center}
   \caption{Visualization of detection results from different models on negative images for some descriptions. There is no GT instance on these images for the descriptions.
   From left to right: GT, predictions from OVD, REC, bi-functional, and DOD methods.
   Best viewed in color and zoomed in.
   % \vspace{-8pt}
   }
\label{fig:vis_rejection}
\end{figure}

\subsection{Additional evaluation results for DOD}

\noindent \textbf{More comparison between baselines.}
In \cref{tab:comparison_supp} we show a more complete comparison of the evaluated baselines on \ddd{} with different metrics. Results on average recalls are added.
In REC datasets like RefCOCO~\cite{yu2016modeling,mao2016generation}, the standard metric is accuracy (which equals to precision and also recall in REC setting). This is not suitable for DOD, which is essentially a detection task. Here we also report the average recall metric in COCO API, but it does not necessarily correspond to the effectiveness of a method for DOD, which requires rejecting negative instances while REC does not.

As shown in \cref{tab:comparison_supp}, REC methods are bad at recall, possibly because it can only predict one instance for one description, no matter how many instances actually exists in GT.
OVD methods are also bad at this metric though they produce a dozen of output (see \cref{fig:vis_absence,fig:vis_rejection}. This may partially explains its low mAP.
The bi-functional methods and the DOD one are all strong on this metric.
Grounding-DINO, though performs not as good as the proposed OFA-DOD in terms of mAPs, obtains the best recall. This indicates that it tends to produce more detection results.

\noindent \textbf{Inference of bi-functional methods.}
As discussed in Section 5.1 of the main paper, bi-functional methods obtain a 100\% No-instance FPPC and fail to reject negative images on \ddd{}. This is due to the inference strategy based on REC. It is possible to apply other inference strategy for them.

We verify the effect of inference strategy on these two bi-functional methods~\cite{yan2023universal,liu2023groundingdino}, with No-instance FPPC and overall \textsl{FULL} mAP, and make comparison with the proposed baseline. As shown in \cref{tab:binfunctional_thres}, we try to apply a threshold to filter out certain low-score predictions, similar to the post-processing steps in OVD~\cite{minderer2022simpleOWLViT}.
With this inference strategy, we observe that the increase of score threshold does lower the No-instance FPPC significantly, but at the cost of overall mAP. Therefore, we apply the REC-based inference strategy for these bi-functional methods by default.

Furthermore, we find that when the score threshold is quite high (0.7 for Grounding-DINO and 0.8 for UNINEXT), they reach a FPPC similar to the proposed baseline but with much less overall mAP (15.7 mAP for UNINEXT and 13.6 mAP for Grounding-DINO, while ours 21.6 mAP). Therefore, it might be fair to say that the proposed baseline achieves a better balance between the ability to reject negative images and the overall detection capability.

% \Todo{mAP (or other overlap metric) to show the difference w/ or wo/ the absence word like "no".}

% \subsection{The proposed baseline on OVD and REC}

% We evaluate OFA-DOD on OVD/REC datasets. The results indicate \textbf{substantial improvements over OFA} for both OVD and REC. This shows the improvements of OFA-DOD over OFA make it better for DOD, OVD and REC, and when a model is improved to be more suitable for DOD, it also exhibit corresponding performance gains on REC and OVD. This implies that the DOD task is compatible with REC and OVD.

% Compared with SOTAs on REC (without fine-tuning), OFA-DOD outperforms the SOTA Grounding-DINO. Compared with SOTAs on OVD (fine-tuning on base classes), OFA-DOD is not good as CORA~\cite{wu2023cora}, but outperforms Detic~\cite{zhou2022detecting} on novel classes, showing good generalization ability. We argue the reason why OFA-DOD does not obtain SOTA on OVD is: the original OFA is not suitable for detection tasks, incapable of rejecting negative instances, lacking compatibility with multi-target outputs and yielding poor results. Although OFA-DOD has augmented its detection ability and improved its performance on OVD by more than 20 mAP, it is still far from perfect for OVD and DOD. This is no surprise as it is only a baseline for future research.

\begin{figure}[t]
\begin{center}
   \includegraphics[width=\linewidth]{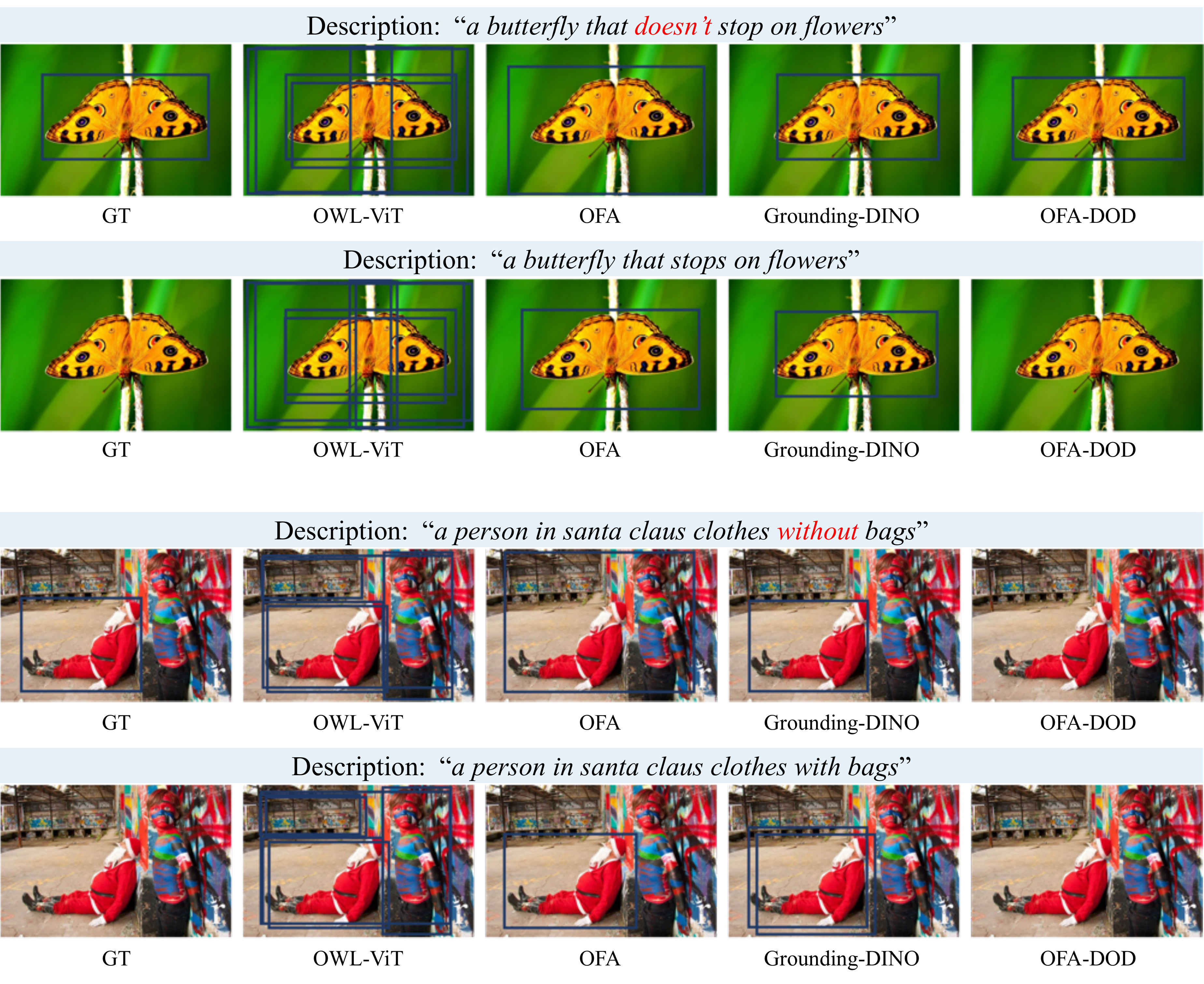}
    % \vspace{-25pt}
\end{center}
   \caption{Visualization of detection results from different models on absence descriptions and their contradictory presence descriptions.
   The key words in absence descriptions are highlighted in red.
   From left to right: GT, predictions from OVD, REC, bi-functional, and DOD methods.
   Best viewed in color and zoomed in.
   % \vspace{-8pt}
   }
\label{fig:vis_absence}
\end{figure}

\subsection{Visual comparisons}

% \noindent \textbf{Images with multiple instances.}

\noindent \textbf{Rejecting negative samples.}
As shown in \cref{fig:vis_rejection}, we visualized two descriptions and two images with no corresponding GT instance. An ideal DOD method should refrain from predicting instances.
OWL-ViT~\cite{minderer2022simpleOWLViT}, the OVD method, predicts multiple instances on these images, some of which overlap with each other. Such redundant predictions are not suitable for this setting.
OFA~\cite{wang2022ofa}, the REC method, always predicts an instance for one reference, making it highly prone to mistakes in such negative images.
Grounding-DINO~\cite{liu2023groundingdino}, the bi-functional method, correctly locates the \texttt{hot air balloon} and \texttt{dog} but fails to capture features related to \texttt{with words} and \texttt{clothed} in the language description.
In the last row, the proposed baseline for DOD successfully rejects one negative image but fails with the other one. This implies that it may perform better on such challenges compared to previous methods, but is still far from being strong.

\noindent \textbf{Absence or presence descriptions.}
In \cref{fig:vis_absence}, we present the detection results for two pairs of descriptions, each with one absence description and its exact counterpart presence description. We visualize the GT (Ground Truth) and also predictions from 4 representative methods.

In the first pair, \texttt{a butterfly that \textcolor{red}{doesn't} stop on flowers}, the GT exists for the absence description, but not for the corresponding presence counterpart. We observe that previous methods are not sensitive to the distinction between presence and absence, leading to similar results for both descriptions. However, the proposed baseline stands as an exception by correctly predicting the bounding box for the absence description and successfully rejecting the presence one. This could be attributed to the language comprehension ability of OFA, as it is trained on multiple text-related tasks.

In the second pair, \texttt{a person in santa claus clothes \textcolor{red}{without} bags}, most methods also yield similar results for both descriptions. Although OFA produces noticeably different bounding boxes for two descriptions, the one corresponding to the absence description is overly large, while the one for the presence description results in a negative prediction. Unfortunately, the proposed baseline incorrectly rejects the predictions for this case.

\end{document}